\def\year{2021}\relax
\documentclass[letterpaper]{article} 
\usepackage{aaai21}  
\usepackage{times}  
\usepackage{helvet} 
\usepackage{courier}  
\usepackage[hyphens]{url}  
\usepackage{graphicx} 
\usepackage{natbib}  
\usepackage{caption} 
\usepackage[switch]{lineno}

\usepackage{comment}
\usepackage{amsmath,amssymb} 
\usepackage{color}
\usepackage{capt-of}
\usepackage{soul}
\usepackage{multirow}
\usepackage{bbold}

\usepackage{stmaryrd}
\usepackage{subfigure}
\usepackage{booktabs} 
\usepackage{amsmath}
\usepackage{amssymb}
\usepackage{algorithm}
\usepackage{algorithmicx}
\usepackage[noend]{algpseudocode}

\usepackage{makecell}

\newtheorem{definition}{Definition}

\newcommand{\eg}{\textit{e}.\textit{g}.}

\title{Visualization of Supervised and Self-Supervised Neural Networks\\ via Attribution Guided Factorization}

\author{Shir Gur, Ameen Ali, Lior Wolf
}
\affiliations{

    The School of Computer Science, Tel Aviv University, Tel Aviv, Israel
}

\begin{document}

\maketitle

\begin{abstract}
Neural network visualization techniques mark image locations by their relevancy to the network's classification. Existing methods are effective in highlighting the regions that affect the resulting classification the most. However, as we show, these methods are limited in their ability to identify the support for alternative classifications, an effect we name {\em the saliency bias} hypothesis. In this work, we integrate two lines of research: gradient-based methods and attribution-based methods, and develop an algorithm that provides per-class explainability.
The algorithm back-projects the per pixel local influence, in a manner that is guided by the local attributions, while correcting for salient features that would otherwise bias the explanation.
In an extensive battery of experiments, we demonstrate the ability of our methods to class-specific visualization, and not just the predicted label. Remarkably, the method obtains state of the art results in benchmarks that are commonly applied to gradient-based methods as well as in those that are employed mostly for evaluating attribution methods. Using a new unsupervised procedure, our method is also successful in demonstrating that self-supervised methods learn semantic information.
Our code is available at: \url{https://github.com/shirgur/AGFVisualization}.
\end{abstract}

\section{Introduction}
\label{sec:intro}
\begin{figure*}[t]
    \setlength{\tabcolsep}{4.5pt} 
    \centering
    \begin{tabular}{cccccc}
        \includegraphics[width=0.14\linewidth]{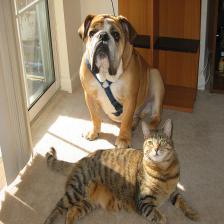} &
        \includegraphics[width=0.14\linewidth]{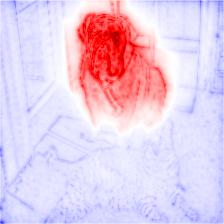} &
        \includegraphics[width=0.14\linewidth]{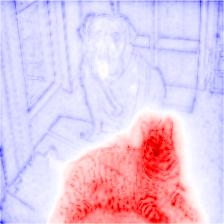} &
        \includegraphics[width=0.14\linewidth]{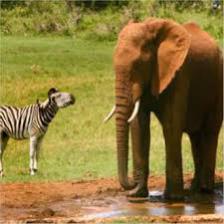} &
        \includegraphics[width=0.14\linewidth]{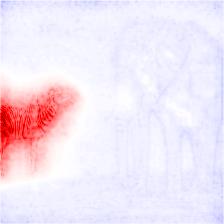} &
        \includegraphics[width=0.14\linewidth]{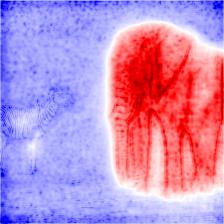}\\
        (a) & (b) & (c) & (d) & (e) & (f)
    \end{tabular}
    \caption{Visualizations by our method for a pre-trained VGG-19. (a,d) input images. (b,e) the heatmap generated for the top label. (c,f) same for the 2nd highest prediction.}
    \label{fig:title}
\end{figure*}
The most common class of explainability methods for image classifiers visualize the reason behind the classification of the network as a heatmap. These methods can make the rationale of the decision accessible to humans, leading to higher confidence in the ability of the classifier to focus on the relevant parts of the image, and not on spurious associations, and help debug the model. In addition to the human user, the ``computer user'' can also benefit from such methods, which can seed image segmentation techniques~\citep{ahn2019weakly,huang2018weakly,wang2019self,hoyer2019grid}, or help focus generative image models, among other tasks.

The prominent methods in the field can be divided into two families: (i) gradient-based maps, which consider the gradient signal as it is computed by the conventional backpropagation approach~\citep{sundararajan2017axiomatic,smilkov2017smoothgrad,srinivas2019full,selvaraju2017grad} and (ii) relevance propagation methods~\citep{bach2015pixel,nam2019relative,gu2018understanding,iwana2019explaining}, which project high-level activations back to the input domain, mostly based on the deep Taylor decomposition by~\cite{montavon2017explaining}. The two families are used for different purposes, and are evaluated by different sets of experiments and performance metrics. As we show in Sec.~\ref{sec:exp}, both types of methods have complementary sets of advantages and disadvantages: gradient based methods, such as Grad-CAM, are able to provide a class specific visualization for deep layers, but fail to do so for the input image, and also provide a unilateral result. In contrast, attribution based methods, excel in visualizing at the input image level, and have bipartite results, but lack in visualizing class specific explanations.

We present a novel method for class specific visualization of deep image recognition models. The method is able to overcome the limitations of the previous work, by combining ideas from both families of methods, and accumulating across the layers both gradient information, and relevance attribution. The method corrects for what we term {\em the saliency bias}.  This bias draws the attention of the network towards the salient activations, and can prevent visualizing other image objects. 
This has led to the claims that visualization methods mimic the behavior of edge detectors, and that the generated heatmaps are largely independent of the network weights~\citep{adebayo2018sanity}.

There are two different questions that explainability method often tackle: (i) which pixels affect classification the most, and (ii) which pixels are identified as belonging to the predicted class. Our method answers the second one and outperforms the relevant literature methods in multiple ways.
First, the locations we identify are much more important to the classification outcome than those of the baselines, when looking inside the region of a target class. Second, we are able to identify regions of multiple classes in each image, and not just the prominent class, see Fig.~\ref{fig:title}. Our method greatly outperforms recent  multi-class work~\citep{gu2018understanding,iwana2019explaining}.

The main contributions of our work are: 
(1) a novel explainability method, that combines both gradient and attribution techniques, as well as a new attribution guided factorization technique for extracting informative class-specific attributions from the input feature map and its gradients. 
(2) we point to and correct, for the first time, for \textit{the saliency bias}. This bias hindered all previous attempts to explain more than the top decision, and is the underlying reason for the failures of explainability methods demonstrated in the literature.
(3) state-of-the-art performance in both negative perturbation and in segmentation-based evaluation. The former is often used to evaluate attribution methods, and the latter is often used to evaluate gradient-based methods. 
(4) using a novel procedure, we show for the first time as far as we can ascertain, that self-supervised networks implicitly learn semantic segmentation information.

\section{Related work}
Many explainability methods belong to one of two classes: attribution and gradient methods. Both aim to explain and visualize neural networks, but differ in their underlying concepts. A summary of these methods with their individual properties is presented in Tab.~\ref{tab:methods}.

Methods outside these two classes include those that generate salient feature maps~\cite{dabkowski2017real,simonyan2013deep,mahendran2016visualizing,zhou2016learning,zeiler2014visualizing,zhou2018interpreting}, Activation Maximization~\cite{erhan2009visualizing} and Excitation Backprop~\cite{zhang2018top}. 
Extremal Perturbation methods~\cite{fong2019understanding,fong2017interpretable} are applicable to black box models, but suffer from high computational complexity. Shapley-value based methods~\cite{lundberg2017unified}, despite their theoretical appeal, are known to perform poorly in practice. Therefore, while we compare empirically with several Shaply and perturbation methods, we focus on the the newer gradient and attribution methods.

{\bf Attribution propagation} methods follow the Deep Taylor Decomposition (DTD) method, of~\citet{montavon2017explaining}, which decompose the network classification decision into the contributions of its input elements. Following this line of work, methods, such as Layer-wise Relevance Propagation (LRP)by \citet{bach2015pixel}, use DTD to propagate relevance from the predicated class, backward, to the input image, in neural networks with a rectified linear unit (ReLU) non-linearity. The PatterNet and PaternAttribution~\cite{kindermans2017learning} methods yield similar results to LRP. A disadvantage of LRP, is that it is class agnostic, meaning that propagating from different classes yields the same visualization. Contrastive-LRP (CLRP) by~\citet{gu2018understanding} and Softmax-Gradient-LRP (SGLRP) by~\citet{iwana2019explaining} use LRP to propagate results of the target class and, in contrast to all other classes, in order to produce a class specific visualization. \citet{nam2019relative} presented RAP, a DTD approach that partitions attributions to positive and negative influences, following a mean-shift approach. This approach is class agnostic, as we demonstrate in Sec.~\ref{sec:exp}. Deep Learning Important
FeaTures (DeepLIFT)~\cite{shrikumar2017learning} decomposes the output prediction, by assigning the differences of contribution scores between the activation of each neuron to its reference activation.

{\bf Gradient based} methods use backpropagation to compute the gradients with respect to the layer's input feature map, using the chain rule. The Gradient*Input method by~\citet{shrikumar2016not} computes the (signed) partial derivatives of the output with respect to the input, multiplying it by the input itself. Integrated Gradients~\cite{sundararajan2017axiomatic}, similar to~\citet{shrikumar2016not}, computes the multiplication of the inputs with its derivatives, only they compute the average gradient while performing a linear interpolation of the input, according to some baseline that is defined by the user. 
SmoothGrad by~\citet{smilkov2017smoothgrad}, visualize the mean gradients of the input, while adding to the input image a random Gaussian noise at each iteration.
The FullGrad method by~\citet{srinivas2019full} suggests computing the gradients of each layer's input, or bias, followed by a post-processing operator, usually consisting of the absolute value, reshaping to the size of the input, and summing over the entire network. Where for the input layer, they also multiply by the input image before post-processing. As we show in Sec.~\ref{sec:exp}, FullGrad produces class agnostic visualizations. On the other hand, Grad-CAM by~\citet{selvaraju2017grad} is a class specific approach, combining both the input features, and the gradients of a network's layer. This approach is commonly used in many applications due to this property, but its disadvantage rests in the fact that it can only produce results for very deep layers, resulting in coarse visualization due to the low spatial dimension of such deep layers.

\begin{table*}[t]
\begin{tabular*}{\linewidth}{@{\extracolsep{\fill}}l@{~~}c@{~~}c@{~~}c@{~~}c@{~~}c@{~~}c@{~~}c@{~~}c@{~~}c@{~~}c@{}}
\toprule
 & Int. & Smooth & \multirow{2}{*}{LRP} & \multirow{2}{*}{LRP$_{\alpha \beta}$} & \multirow{1}{*}{Full} & Grad & \multirow{2}{*}{RAP} & \multirow{2}{*}{CLRP} & \multirow{2}{*}{SGLRP} & \multirow{2}{*}{Ours} \\
 & Grad & Grad & & & Grad & CAM &  & & &  \\
 \midrule
Gradients & \checkmark & \checkmark & & & \checkmark & \checkmark & & & & \checkmark\\
Attribution & & & \checkmark & \checkmark & & & \checkmark & \checkmark & \checkmark & \checkmark\\
Class Specific &  & & & &  & \checkmark & & \checkmark & \checkmark & \checkmark\\
Input Domain & \checkmark & \checkmark & \checkmark & \checkmark & \checkmark & & \checkmark & \checkmark & \checkmark & \checkmark\\
\bottomrule
\end{tabular*}
\caption{Properties of various visualization methods, which can be largely divided into gradient based and attribution based. Most methods are not class specific, and except for Grad-CAM, all methods project all the way back to the input image.}
\label{tab:methods}
\end{table*}

\section{Propagation methods}
\label{sec:preliminaries}

We define the building blocks of attribution propagation and gradient propagation that are used in our method.

\noindent\textbf{Attribution Propagation:}
Let $x^{(n)}$, $\theta^{(n)}$ be the input feature map and weights of layer $L^{(n)}$, respectively, where $n\in [1\dots N]$ is the layer index in a network, consisting of $N$ layers. In this field's terminology, layer $n-1$ is downstream of layer $n$, layer $N$ processes the input, while layer $1$ produces the final output.

Let $x^{(n-1)}=L^{(n)}(x^{(n)}, \theta^{(n)})$ to be the result of applying layer $L^{(n)}$ on $x^{(n)}$. The relevancy of layer $L^{(n)}$ is given by $R^{(n)}$ and is also known as the attribution.
\begin{definition}{\em\cite{montavon2017explaining}}
\label{def:generic_rel}
The {\em generic attribution propagation rule} is defined, for two tensors, $\mathbf{X}$ and $\Theta$, as:
\begin{align}
    \label{eq:generic_rel}
    R^{(n)}_j &= \mathcal{G}_j^{(n)}(\mathbf{X}, \Theta, R^{(n-1)})  \\\nonumber
    &= \sum_i \mathbf{X}_j\frac{\partial L^{(n)}_i(\mathbf{X}, \Theta)}{\partial \mathbf{X}_j} \frac{R^{(n-1)}_i}{\sum_{i'}L^{(n)}_{i'}(\mathbf{X}, \Theta)}
\end{align}
\end{definition}
Typically, $\mathbf{X}$ is related to the layer's input $x^{(n)}$, and $\Theta$ to its weights $\theta^{(n)}$. LRP~\cite{binder2016layer} can be written in this notation by setting $\mathbf{X} = {x}^{(n)+}$ and $\Theta = {\theta}^{(n)+}$, where $\tau^+ = \max(0, \tau)$ for a tensor $\tau$. 
{Note that Def.~\ref{def:generic_rel} satisfies the {\em conservation rule}:}
\begin{align}
    \label{eq:conservation_rule}
    \sum_j R^{(n)}_j = \sum_i R^{(n-1)}_i
\end{align}

Let $y$ be the output vector of a classification network with $\mathcal{C}$ classes, and let $y^t$ represent the specific value of class $t \in \mathcal{C}$. LRP defines $R^{(0)} \in \mathbb{R}^{|\mathcal{C}|}$ to be the a zeros vector, except for index $t$, where $R_t^{(0)} = y^t$. 
Similarly, the CLRP~\cite{gu2018understanding} and SGLRP~\cite{iwana2019explaining} methods calculate the difference between two LRP results, initialized with two opposing $R^{(0)}$ for ``target'' and ``rest'', propagating relevance from the ``target'' class, and the ``rest'' of the classes, \eg CLRP is defined as:
\begin{align}
    CLRP &= R^{(N)}_{tgt} - \mathcal{N}\big(R^{(N)}_{rst}, R^{(N)}_{tgt}\big)
\end{align}
\noindent
where $R^{(0)}_{tgt} = R^{(0)}$, $R^{(0)}_{rst} = (y-R^{(0)})/(|\mathcal{C}|-1)$, and  $\mathcal{N}$ is a normalization term $\mathcal{N}(a, b) = a \frac{\sum b}{\sum a}$.

\noindent\textbf{$\Delta$-Shift:}
Def.~\ref{def:generic_rel} presented a generic propagation rule that satisfies the conservation rule in Eq.~\ref{eq:conservation_rule}. However, in many cases, we would like to add a residual signal denoting another type of attribution. The $\Delta$-Shift corrects for the deviation from the conservation rule. 

\begin{definition}
\label{def:delta_shift}
Given a generic propagation result $\mathcal{G}^{(n)}$, following Eq.~\ref{eq:conservation_rule}, and a residual tensor {$\mathbf{r}^{(n)}$}, the $\Delta$-Shift is defined as follows:
\begin{align}
    \Delta^{(n)}_\text{\em shift}(\mathcal{G}^{(n)}, \mathbf{r}^{(n)}) = \mathcal{G}^{(n)} + \mathbf{r}^{(n)}- \frac{\sum \mathbf{r}^{(n)}}{\sum \mathbb{1}_{\mathcal{G}^{(n)} \ne 0}}
\end{align}

\end{definition}
Note that we divide the sum of the residual signal by the number of non-zero neurons.  
While not formulated this way, the RAP method~\cite{nam2019relative} employs this type of correction defined in Def.~\ref{def:delta_shift}.

\noindent\textbf{Gradient Propagation:}
The propagation in a neural network is defined by the chain rule.
\begin{definition}
Let $\mathcal{L}$ be the loss of a neural network. The input feature gradients, $x^{(n)}$ of layer $L^{(n)}$, with respect to $\mathcal{L}$ are defined by the chain rule as follows:
\begin{align}
    \label{eq:chain_rule}
    \nabla x_j^{(n)} := \frac{\partial \mathcal{L}}{\partial x_j^{(n)}} &= \sum_{i} \frac{\partial \mathcal{L}}{\partial x_i^{(n-1)}} \frac{\partial x_i^{(n-1)}}{\partial x_j^{(n)}}
\end{align}
\label{def:chain_rule}
\end{definition} 
Methods such as FullGrad~\cite{srinivas2019full} and SmoothGrad~\cite{smilkov2017smoothgrad} use the raw gradients, as defined in Eq.\ref{eq:chain_rule}, for visualization. Grad-CAM~\cite{selvaraju2017grad}, on the other hand, performs a weighted combination of the input feature gradients, in order to obtain a class specific visualization, defined as follows:
\begin{align}
    \label{eq:grad-cam}
    \text{Grad-CAM}\big(x^{(n)},\nabla x^{(n)}\big) &= \bigg(\frac{1}{|C|}\sum_{c \in [C]} x_c^{(n)}\sum\limits_{\substack{h \in [H]\\w \in [W]}}\nabla x_{c,h,w}^{(n)}\bigg)^+
\end{align}
where $\nabla x_{c,h,w}^{(n)}$ is the specific value of the gradient $C$-channel tensor $x^{(n)}$ at channel $c$ and pixel $(h,w)$, and $x_c^{(n)}$ is the entire channel, which is a matrix of size $H\times W$. 

\noindent\textbf{Guided Factorization:}
The explanation should create a clear separation between the positively contributing regions, or the foreground, and the negatively contributing ones, referred to as the background. This is true for both the activations and the gradients during propagation. Ideally, the relevant data would be partitioned into two clusters --- one for positive contributions and one for the negative contributions.  We follow the partition problem~\cite{yuan2015factorization,gao2016factorization}, in which the data is divided spatially between positive and negative locations, in accordance with the sign of a partition map $\phi \in \mathbb{R}^{H \times W}$.

Given a tensor $\mathbf{Y}\in\mathbb{R}^{C \times H \times W}$,  we re-write it as a matrix in the form of $\mathbf{Y}\in\mathbb{R}^{C \times H W}$. We compute the Heaviside function of $\mathbf{Y}$ using a $\operatorname{sigmoid}$ function:
$\mathbf{H} = \operatorname{sigmoid}(\mathbf{Y})$.
The matrix $H \in [0, 1]^{C \times HW}$ is a positive-matrix, and we consider the following two-class non-negative matrix factorization $\mathbf{H} = \mathbf{R}\mathbf{W}$, 
where $\mathbf{W} \in {\mathbb{R}^+}^{2 \times HW}$ contains the spatial mixing weights, and the representative matrix $\mathbf{R} = [\mathbf{R}_b \, \mathbf{R}_f]$, defined by the mean of each class in the data tensor $\mathbf{H}$ based on the assignment of $\phi$:
\begin{align}
    {\mathbf{R}^f_c} &= \frac{\sum\limits_i^{HW}\mathbf{H}(\mathbf{Y})_{c,i} \odot \mathbb{1}_{\phi_i > 0}}{\sum\limits_i^{HW} \mathbb{1}_{\phi_i > 0}} \,, 
    {\mathbf{R}^b_c} &= \frac{\sum\limits_i^{HW}\mathbf{H}(\mathbf{Y})_{c,i} \odot \mathbb{1}_{\phi_i \leqslant 0}}{\sum\limits_i^{HW} \mathbb{1}_{\phi_i \leqslant 0}}
    \nonumber 
\end{align}
where $\mathbf{R}^f,\mathbf{R}^b \in {\mathbb{R}^+}^C$, $c\in C$ is the channel dimension and $\odot$ denotes the Hadamard product.

We estimate the matrix $\mathbf{W}$ of positive weights by least squares $\mathbf{W} = [\mathbf{W}_b \, \mathbf{W}_f] = ((\mathbf{R}^\top\mathbf{R})^{-1}\mathbf{R}^\top \mathbf{H})^+$, 
where $\mathbf{W}_f,\mathbf{W}_b \in {\mathbb{R}^+}^{HW}$. Combining the foreground weights $W_f$ with the background weights $W_b$ into the same axis is done by using both negative and positive values, leading to the following operator:
$\mathcal{F}(\mathbf{Y},\phi)= \mathbf{W}_f - \mathbf{W}_b$,
where $\mathcal{F}$ is a function that takes $\mathbf{Y}$ and $\phi$, as inputs. 
We further normalize $\mathbf{Y}$ using $\mathcal{N}_{\max}(a) = \frac{a}{\max(a)}$, to allow multiple streams to be integrated together:
\begin{align}
    \label{def:norm_fact}
    \bar{\mathcal{F}}(\mathbf{Y}, \phi) = \mathcal{F}(\mathcal{N}_{\max}(\mathbf{Y}), \phi)
\end{align}

\section{The Integrated Method}

\label{sec:method}

Let $M$ be a multiclass CNN classifier ($\mathcal{C}$ labels), and $I=x^{(N)}$ be the input image. The network $M$ outputs a score vector $y \in \mathbb{R}^{|\mathcal{C}|}$, obtained before applying the $\operatorname{softmax}$ operator.
Given any target class $t$, our goal is to explain where (spatially) in $I$ lies the support for class $t$. The method is composed of two streams, gradients and attribution propagation. Each step, we use the previous values of the two, and compute the current layer's input gradient and attribution.

There are three major components to our method, (i) propagating attribution results using Def.~\ref{def:generic_rel}, (ii) factorizing the activations and the gradients in a manner that is guided by the attribution, and (iii) performing attribution aggregation and shifting the values, such that the conservation rule is preserved. 
The shift splits the neurons into those with a positive and negative attribution. The complete algorithm is listed as part of the supplementary material.

\subsection{Initial Attribution Propagation}
As shown in~\cite{gu2018understanding,iwana2019explaining}, the use of different initial relevance can result in improved results. 
Let $\Phi^{(n)}$ and $\Phi^{(n-1)}$ be the output and input class attribution maps of layer $L^{(n)}$, respectively. We employ the following initial attribution for explaining decision $t$. Let $y:=x^{(0)}$ be the  output vector of the classification network (logits), we compute the initial attribution $\Phi^{(1)}$ as:
\begin{multline}
    \hat{y} = \operatorname{softmax} \big(y^t\exp {\big(- \frac{1}{2}\big(\frac{y-y^t}{\max\|y-y^t\|_1}\big)^2\big)}\big) \\
    \frac{\partial \hat{y}^t}{\partial x^{(1)}_j} = \sum_{i} \frac{\partial \hat{y}^t}{\partial y_i} \frac{\partial y_i}{\partial x_j^{(1)}}\,,
    \Phi^{(1)} = x^{(1)} \odot \frac{\partial \hat{y}^t}{\partial x^{(1)}}
\end{multline}

\noindent In this formulation, we replace the psuedo-probabilities of vector $y$ with another vector, in which class $t$ that we wish to provide an explanation for is highlighted, and the rest of the classes are scored by the closeness of their assigned probability to that of $t$. This way, the explanation is no longer dominated by the predicted class.

\begin{figure*}[t]
    \setlength{\tabcolsep}{3pt} 
    \centering
    \begin{tabular}{ccccccccc}
         &Input &  \multirow{2}{*}{LRP} & \multirow{2}{*}{FullGrad} & Grad & \multirow{2}{*}{RAP} & \multirow{2}{*}{CLRP} & \multirow{2}{*}{SGLRP} & \multirow{2}{*}{Ours} \\
         &Image &  & & CAM &  & & &  \\
        (a) &
        \includegraphics[width=0.103\textwidth]{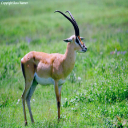} &
        \includegraphics[width=0.103\textwidth]{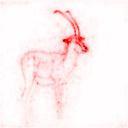} &
        \includegraphics[width=0.103\textwidth]{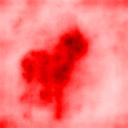} &
        \includegraphics[width=0.103\textwidth]{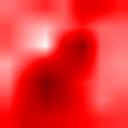} &
        \includegraphics[width=0.103\textwidth]{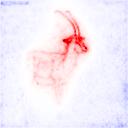} &
        \includegraphics[width=0.103\textwidth]{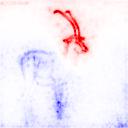} &
        \includegraphics[width=0.103\textwidth]{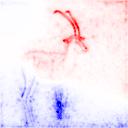} &
        \includegraphics[width=0.103\textwidth]{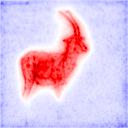}\\
        \hline
        (b) &
        \includegraphics[width=0.103\textwidth]{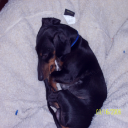} &
        \includegraphics[width=0.103\textwidth]{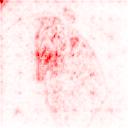} &
        \includegraphics[width=0.103\textwidth]{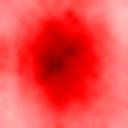} &
        \includegraphics[width=0.103\textwidth]{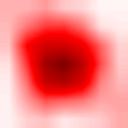} &
        \includegraphics[width=0.103\textwidth]{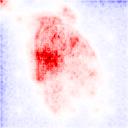} &
        \includegraphics[width=0.103\textwidth]{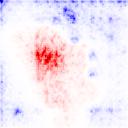} &
        \includegraphics[width=0.103\textwidth]{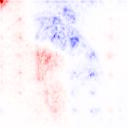} &
        \includegraphics[width=0.103\textwidth]{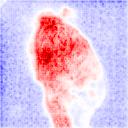}\\
        \hline
        \multirow{2}{*}{(c)} &
        \raisebox{8mm}{\multirow{2}{*}{\makecell{Dog $\rightarrow$\\\includegraphics[width=0.103\textwidth]{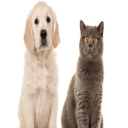}\\ Cat $\rightarrow$}}} &
        \includegraphics[width=0.103\textwidth]{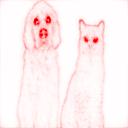} &
        \includegraphics[width=0.103\textwidth]{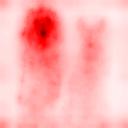} &
        \includegraphics[width=0.103\textwidth]{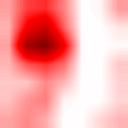} &
        \includegraphics[width=0.103\textwidth]{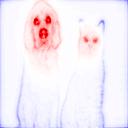} &
        \includegraphics[width=0.103\textwidth]{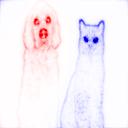} &
        \includegraphics[width=0.103\textwidth]{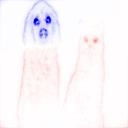} &
        \includegraphics[width=0.103\textwidth]{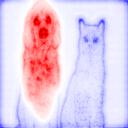} \\
        &&
        \includegraphics[width=0.103\textwidth]{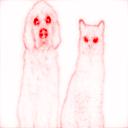} &
        \includegraphics[width=0.103\textwidth]{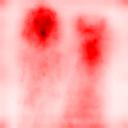} &
        \includegraphics[width=0.103\textwidth]{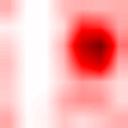} &
        \includegraphics[width=0.103\textwidth]{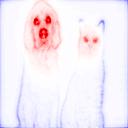} &
        \includegraphics[width=0.103\textwidth]{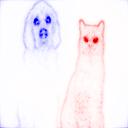} &
        \includegraphics[width=0.103\textwidth]{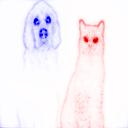} &
        \includegraphics[width=0.103\textwidth]{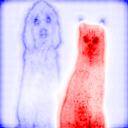}
        
    \end{tabular}
    \caption{(a) A comparison between methods for VGG-19. (b) Same for ResNet-50. (c) Visualization of two different classes for VGG-19. {Many more results can be found in the supplementary material.}}
    \label{fig:results}
\end{figure*}
\begin{figure*}[t]
    \centering
    \begin{tabular*}{\textwidth}{@{\extracolsep{\fill}}cc}
        \includegraphics[height=0.23\linewidth]{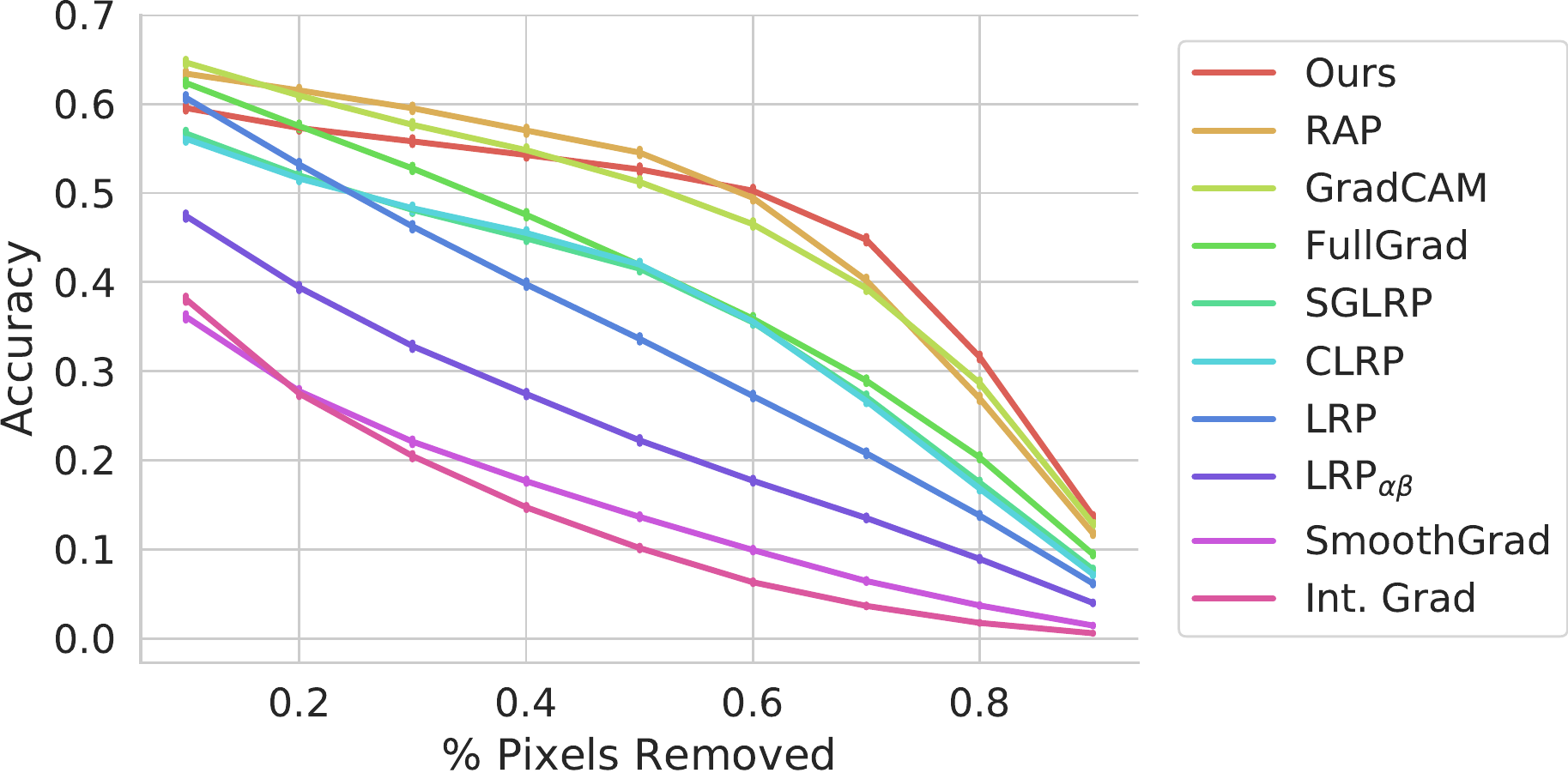} &
        \includegraphics[height=0.23\linewidth]{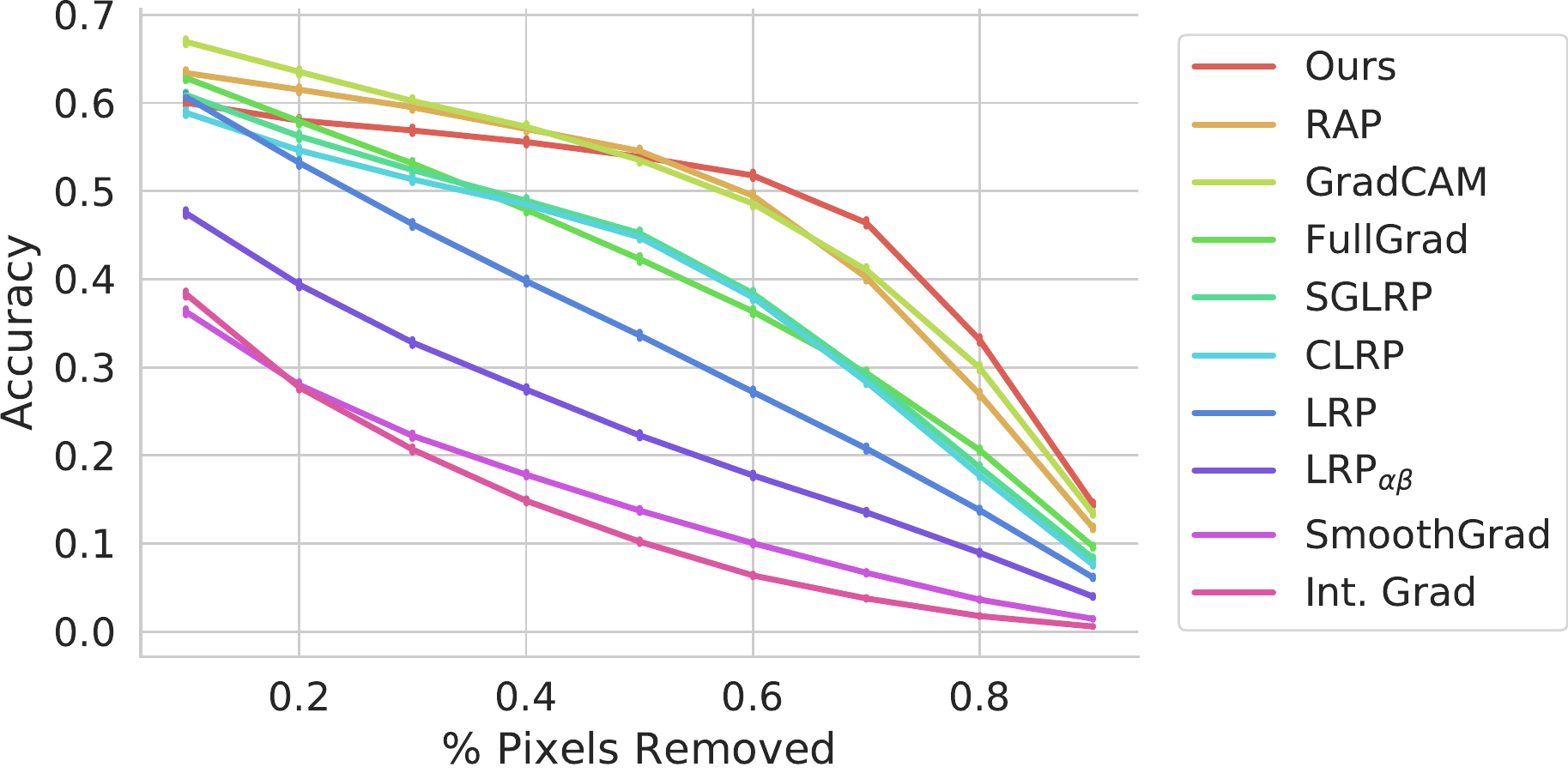}\\
        (a) & (b)\vspace{-3mm}
    \end{tabular*}    
    \caption{{Negative perturbation results on the ImageNet validation set of \textbf{(a)} the predicted and \textbf{(b)} the target class. Show is the change in accuracy, when removing a fraction of the image according to the attribution value, starting from lowest to highest.
    }}
    \label{fig:neg_pert}
\end{figure*}
\begin{table*}[t]
\begin{tabular*}{\textwidth}{@{\extracolsep{\fill}}lcccccccccc}
\toprule
 & Int. & Smooth & \multirow{2}{*}{LRP} & \multirow{2}{*}{LRP$_{\alpha \beta}$} & \multirow{1}{*}{Full} & Grad & \multirow{2}{*}{RAP} & \multirow{2}{*}{CLRP} & \multirow{2}{*}{SGLRP} & \multirow{2}{*}{Ours} \\
 & Grad & Grad & & & Grad & CAM &  & & &  \\
 \midrule
Predicted & 10.4 & 12.0 & 26.8 & 18.8 & 32.1 & 37.8 & 38.7 & 29.8 & 29.9 & \textbf{38.9}\\
Target    & 10.5 & 12.1 & 26.8 & 18.8 & 32.4 & 39.4 & 38.7 & 31.6 & 32.4 & \textbf{40.0}\\
\bottomrule
\end{tabular*}
\caption{Area Under the Curve (AUC) for the two negative perturbation tests, showing results for predicted and target class. The class-agnostic methods either perform worse or experience insignificant change on the target class test.}
\label{tab:neg_per_auc}
\bigskip
\begin{tabular*}{\textwidth}{@{\extracolsep{\fill}}c@{~~~}l@{~~~}c@{~~~}c@{~~~}c@{~~~}c@{~~~}c@{~~~}c@{~~~}c@{~~~}c@{~~~}c@{~~~}c@{~~}c@{~~}c@{}}
\toprule
 &  & {Gradient} & {Meaningful} & Int. & Smooth & \multirow{2}{*}{LRP} & \multirow{2}{*}{LRP$_{\alpha \beta}$} & \multirow{1}{*}{Full} & Grad & \multirow{2}{*}{RAP} & \multirow{2}{*}{CLRP} & \multirow{2}{*}{SGLRP} & \multirow{2}{*}{Ours} \\
 Dataset& Metric & SHAP & Perturbation & Grad & Grad & & & Grad & CAM &  & & &  \\
 
\midrule

\multirow{2}{*}{ImageNet} &Pixel Acc. & 49.9 & 29.6 & 72.7 & 70.1 & 74.9 & 29.6 & 75.8 & 66.5 & 78.8 & 53.3 & 54.5 & \textbf{79.9}\\
& mAP        & 50.0 & 45.4 & 67.4 & 65.1 & 69.9 & 47.5 & 70.6 & 62.4 & 73.7 & 54.6 & 55.1 & \textbf{76.7}\\

\midrule

\multirow{2}{*}{VOC'12} &Pixel Acc. & 9.3 & 10.2 & 70.1 & 69.9 & 66.6 & 70.8 & 26.9 & 52.6 & 72.8 & 72.8 & 73.1 & \textbf{77.5}\\
& mAP        & 35.2 & 22.7 & 34.9 & 34.3 & 40.8 & 37.0 & 19.7 & 41.7 & 39.6 & 37.9 & 32.6 & \textbf{45.9}\\
\bottomrule
\end{tabular*}
\caption{Quantitative segmentation results on (a) ImageNet and (b) PASCAL-VOC 2012.}
\label{tab:segmentation}
\end{table*}
\subsection{Class Attribution Propagation}
As mentioned above, there are two streams in our method. The first propagates the gradients, which are used for the factorization process we discuss in the next section, and the other is responsible for propagating the resulting class attribution maps $\Phi^{(n-1)}$, using DTD-type propagations. This second stream is more involved, and the computation of $\Phi^{(n)}$ has multiple steps.

The first step is to propagate $\Phi^{(n-1)}$ through $L^{(n)}$ following Eq.~\ref{eq:generic_rel}, using two variants. Both variants employ $\Phi^{(n-1)}$ as the tensor to be propagated, but depending on the setting of $\mathbf{X}$ and $\Theta$, result in different outcomes. 
The first variant considers the absolute influence $\mathbf{C}^{(n)}$, defined by:
\begin{align}
    \mathbf{C}^{(n)} = \mathcal{G}^{(n)}(|x^{(n)}|, |\theta^{(n)}|, \Phi^{(n-1)})
\end{align}
The second variant computes the input-agnostic influence $\mathbf{A}^{(n)}$, following Eq.~\ref{eq:generic_rel}:
\begin{align}
    \mathbf{A}^{(n)} = \mathcal{G}^{(n)}(\mathbb{1}, |\theta^{(n)}|, \Phi^{(n-1)})
\end{align}
where $\mathbb{1}$ is an all-ones tensor of the shape of $x^{(n)}$. We choose the input-agnostic propagation because features in shallow layers, such as edges, are more local and less semantic. It, therefore, reduces the sensitivity to texture.

\subsection{Residual Update}
As part of the method, we compute in addition to $\mathbf{C}^{(n)}$, the factorization of both the input feature map of layer $L^{(n)}$ and its gradients.
This branch is defined by the chain rule in Eq.~\ref{eq:chain_rule}, where we now consider $\mathcal{L} = \hat{y}^t$.
The factorization results in \textit{foreground} and \textit{background} partitions, using guidance from $\mathbf{C}^{(n)}$. This partition follows the idea of our attribution properties, where positive values are part of class $t$, and negatives otherwise. We, therefore, employ the following attribution guided factorization (Eq.~\ref{def:norm_fact} with respect to  $x^{(n)}$ and $\nabla x^{(n)}$) as follows:
\begin{equation}
        \mathbf{F}^{(n)}_{x} = \bar{\mathcal{F}}\big(x^{(n)},\mathbf{C}^{(n)}\big)^+\,,\quad \mathbf{F}^{(n)}_{\nabla x} = \bar{\mathcal{F}}\big(\nabla x^{(n)},\mathbf{C}^{(n)}\big)^+
\end{equation}
note that we only consider the positive values of the factorization update, and that the two results are normalized by their maximal value.
Similarly to~\cite{sundararajan2017axiomatic,selvaraju2017grad}, we define the input-gradient interaction:
\begin{align}
    \mathbf{M}^{(n)}_{x\nabla x} = \mathcal{N}_{\max}\bigg(\bigg(\frac{1}{|C|}\sum_{c \in [C]}\big(x_c^{(n)} \odot \nabla x_c^{(n)}\big)\bigg)^+\bigg)
\end{align}
{The residual attribution is then defined by all attributions other than $\mathbf{C}^{(n)}$:}
\begin{align}
    \mathbf{r}^{(n)} = \mathbf{A}^{(n)} + \mathbf{F}^{(n)}_{\nabla x} + \frac{\mathbf{F}^{(n)}_{x} + \mathbf{M}^{(n)}_{x\nabla x}}{1 + \exp(-\mathbf{C}^{(n)})}
\end{align}
We observe that both $\mathbf{F}^{(n)}_{x}$ and $\mathbf{M}^{(n)}_{x\nabla x}$ are affected by the input feature map, resulting in the {\em saliency bias} effect (see Sec.~\ref{sec:intro}). As a result, we penalize their sum according to $\mathbf{C}^{(n)}$,
in a manner that emphasises positive attribution regions.

We note that $\sum_j (\mathbf{C}_j^{(n)} + \mathbf{r}_j^{(n)}) \neq \sum_i \Phi_i^{(n-1)}$, and the residual needs to be compensated for, in order to preserve the conservation rule. Therefore, we perform a $\Delta$-shift as defined in Def.~\ref{def:delta_shift}, 
resulting in the final attribution:
\begin{align}
    \Phi^{(n)} = \Delta^{(n)}_\text{\em shift}(\mathbf{C}^{(n)}, \mathbf{r}^{(n)})
\end{align}

\subsection {Explaining Self-Supervised Learning (SSL)}
SSL is proving to be increasingly powerful and greatly reduces the need for labeled samples. However, no explainability method was applied to verify that these models, which are often based on image augmentations, do not ignore localized image features. 
 
Since no label information is used, we rely on the classifier of the self-supervised task itself, which has nothing to do with the classes of the datasets. We consider for each image, the image that is closest to it in the penultimate layer's activations. We then subtract the logits of the self supervised task of the image to be visualized and its nearest neighbor, to emphasize what is unique to the current image, and then use explainability methods on the predicted class of the self-supervised task. See supplementary for more details of this novel procedure.

\section{Experiments}
\label{sec:exp}

\noindent\textbf{Qualitative Evaluation:}
Fig.~\ref{fig:results}(a,b) present sample visualization on a representative set of images for networks trained on ImageNet, using VGG19 and ResNet-50, respectively. 
In these figures, we visualize the top-predicted class. More results can be found in the supplementary, including the output for the rest of the methods, which are similar to other methods and are removed for brevity.

The preferable visualization quality provided by our method is strikingly evident. One can observe that (i) LRP, FullGrad and Grad-CAM output only positive results, wherein LRP edges are most significant, and in all three, the threshold between the object and background is ambiguous. (ii) CLRP and SGLRP, which apply LRP twice, have volatile outputs. (iii) RAP is the most consistent, other than ours, but falls behind in object coverage and boundaries. (iv) Our method produces relatively complete regions with clear boundaries between positive and negative regions.

In order to test whether each method is class-agnostic or not, we feed the classifier images containing two clearly seen objects, and propagate each object class separately. In Fig.~\ref{fig:results}(c) we present results for a sample image.
As can be seen, LRP, FullGrad and RAP output similar visualizations for both classes. Grad-CAM, on the other hand, clearly shows a coarse region of the target class, but lacks the spatial resolution. CLRP and SGLRP both achieve class separation, and yet, they are highly biased toward image edges, and do not present a clear separation between the object and its background. Our method provides the clearest visualization, which is both highly correlated with the target class, and is less sensitive toward edges. More samples can be found in the supplementary.

\noindent\textbf{Quantitative Experiments:}
We employ two experiment settings that are used in the literature, negative perturbation and segmentation tests.
We evaluate our method using three common datasets: (i) the validation set of ImageNet~\cite{russakovsky2015ImageNet} (ILSVRC) 2012, consisting of 50K images from 1000 classes, (ii) an annotated subset of ImageNet called ImageNet-Segmentation~\cite{guillaumin2014ImageNet}, containing 4,276 images from 445 categories, and (iii) the PASCAL-VOC 2012 dataset, depicting 20 foreground object classes and one background class, and containing 10,582 images for training, 1449 images for validation and 1,456 images for testing.

\noindent\textbf{Negative Perturbation Experiments:}
The negative perturbation test is composed of two stages, first, a pre-trained network is used to generate the visualizations of the ImageNet validation set. In our experiments, we use the VGG-19 architecture, trained on the full ImageNet training set. Second, we mask out an increasing portion of the image, starting from lowest to highest values, determined by the explainability method. At each step, we compute the mean accuracy of the pre-trained network. We repeat this test twice: once for the explanation of the top-1 predicted class, and once for the ground truth class. The results are presented in Fig.~\ref{fig:neg_pert} and Tab.~\ref{tab:neg_per_auc}. As can be seen, our method achieves the best performance across both tests, where the margin is highest when removing $40\%-80\%$ of the pixels.

\begin{table*}[t]
\begin{minipage}[c]{0.48\linewidth}
    \begin{tabular}{@{}cc}
         \includegraphics[height=82px]{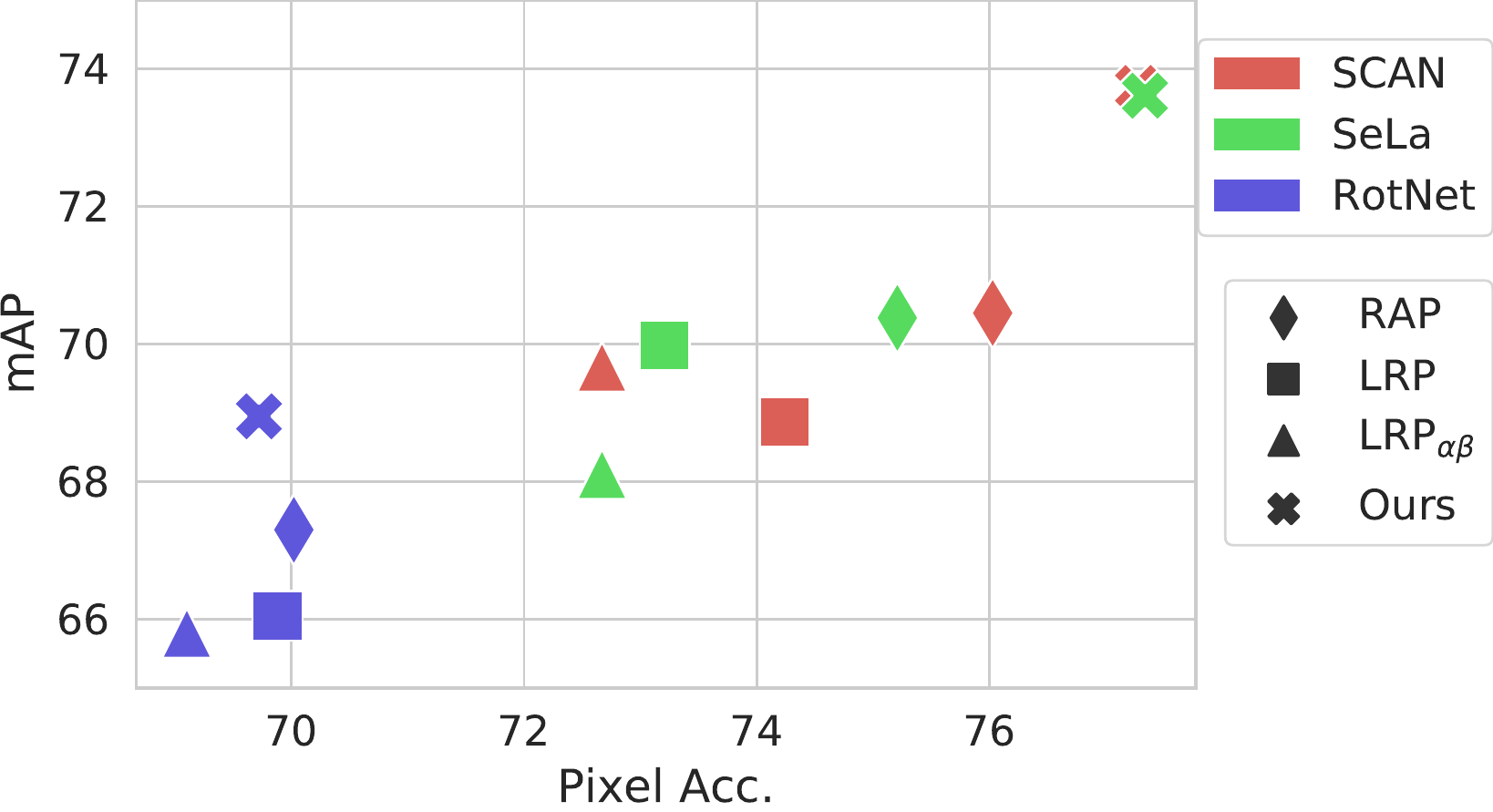}&  \includegraphics[height=82px]{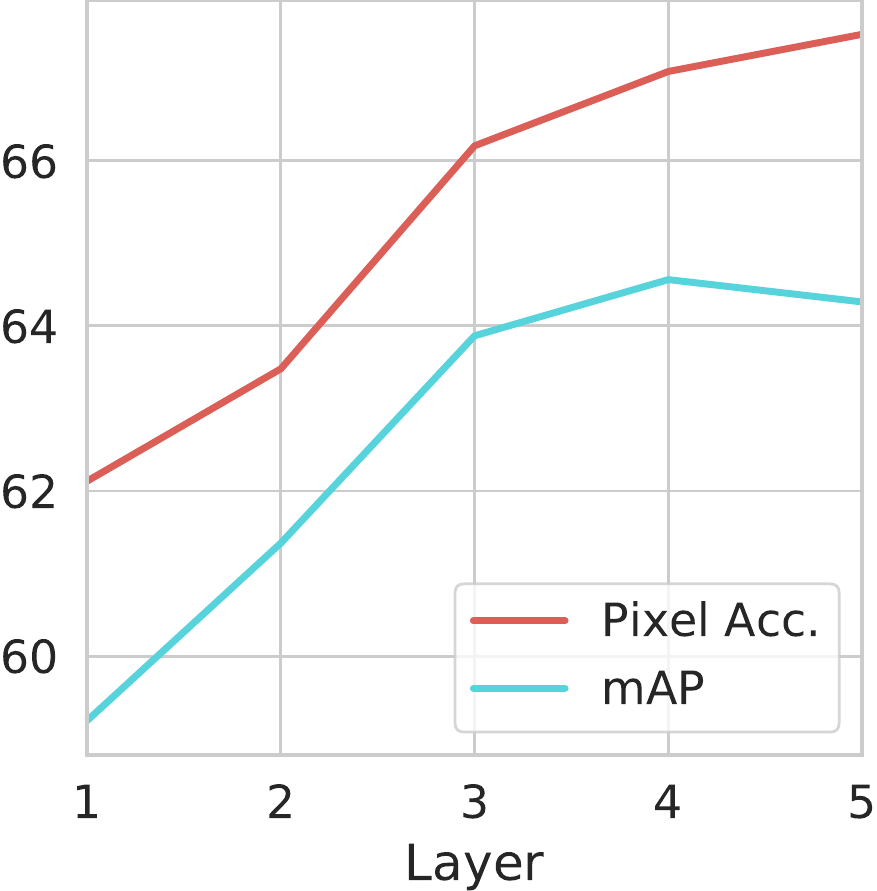}\\
         (a) & (b)
    \end{tabular}
    \captionof{figure}{
    Quantitative results for self-supervised methods in the segmentation task. \textbf{(a)} Comparison of different explainability methods for SCAN, SeLa and RotNet. \textbf{(b)} Per-layer performance of RotNet using linear-probes.
    }
    \label{fig:ssl_fig}
    \bigskip
\end{minipage}%
\hfill
\begin{minipage}[c]{0.48\linewidth}
    \begin{tabular*}{\linewidth}{@{\extracolsep{\fill}}lcccccc}
        \toprule
        & \multicolumn{3}{c}{Ours} & \multicolumn{3}{c}{RAP} \\
        \cmidrule(lr){2-4}
        \cmidrule(lr){5-7}
        Method & ($\Delta$) & ($\textstyle\sum$) & w/o &  ($\Delta$) & ($\textstyle\sum$) & w/o\\
        \midrule
        Supervised & {\bf 42.4}&42.1&41.2&41.6&41.2&40.0\\\
        SeLa & {\bf 37.8} & 37.4 & 36.9 & 37.2 & 36.3& 34.5\\
        SCAN & {\bf 38.1} &37.9&37.0&37.4&36.8&34.8\\
        \bottomrule
    \end{tabular*}
    \caption{
    AUC for negative perturbation tests for self-supervised methods - SeLa and SCAN. ($\Delta$) is our method; ($\textstyle\sum$) is an alternative that adds instead of subtracts, w/o does not consider the neighbor at all. RAP, which is the best baseline in Tab.~\ref{tab:neg_per_auc}, is used as a baseline.
    }
    \label{tab:ssl_tab}
\end{minipage}
\begin{tabular*}{\textwidth}{@{\extracolsep{\fill}}lcccccccc}
\toprule
& \multirow{2}{*}{Ours} & Only & \multicolumn{5}{c}{without} & \multirow{2}{*}{{$\mathbf{r}^{(n)} =$} Grad-CAM}\\
\cline{4-8}
 &  & $\mathbf{C}^{(n)}$ & $\mathbf{A}^{(n)}$ & $\mathbf{F}^{(n)}_x$ & $\mathbf{F}^{(n)}_{\nabla x}$ & $\mathbf{M}^{(n)}$ & $({1 + \exp(-\mathbf{C}^{(n)})})^{-1}$ & \\
 \midrule
AUC & \textbf{38.9} & 38.2 & 34.6 & 37.5 & 37.1 & 37.2 & 37.5 & 37.1 \\
\bottomrule
\end{tabular*}
\caption{AUC results in the negative perturbation test for variations of our method.}
\label{tab:ablation}
\end{table*}
\noindent\textbf{Semantic Segmentation Metrics:}
To evaluate the segmentation quality obtained by each explainability method, we compare each to the ground truth segmentation maps of the ImageNet-Segmentation dataset, and the PASCAL-VOC 2012, evaluating by pixel-accuracy and mean average-precision.
We follow the literature benchmarks for explainability that are based on labeled segments~\cite{nam2019relative}. Note that these are not meant to provide a full weakly-supervised segmentation solution, which is often obtained in an iterative manner. Rather, its goal is to demonstrate the ability of each method without follow-up training.
For the first dataset, we employed the pre-trained VGG19 classifier trained on ImageNet training set, and compute the explanation for the top-predicted class (we have no access to the ground truth class) and compare it to the ground truth mask provided in the dataset.
For the second, we trained a multi-label classifier on the PASCAL-VOC 2012 training set, and consider labels with a probability larger than 0.5 to extract the explainability maps.
For methods that provide both positive and negative values (Gradient SHAP, LRP$\alpha-\beta$, RAP, CLRP, SGLRP, and ours), we consider the positive part as the segmentation map of that object. For methods that provide only positive values (Integrated Grad, Smooth Grad, Full Grad, GradCAM, LRP, Meaningful Perturbation), we threshold the obtained maps at the mean value to obtain the segmentation map. Results are reported in Tab.~\ref{tab:segmentation}, demonstrating a clear advantage of our method over all nine baseline methods, for all datasets and metrics. Other methods seem to work well only in one of the datasets or present a trade-off between the two metrics.

\noindent\textbf{Explainability for  Self-Supervised Models:}
We use three state-of-the-art SSL models: ResNet-50 trained with either SCAN~\cite{vanscan} or SeLa~\cite{asano2019self}, and an Alexnet by~\citet{asano2019critical}, which we denote as RotNet. 

Fig.~\ref{fig:ssl_fig}(a) shows the segmentation performance for the ImageNet ground truth class of the completely unsupervised SSL methods (color) using different explainability methods (shape). For all models our explainability method outperforms the baselines in both mAP and pixel accuracy, except for RotNet where the mAP is considerably better and the pixel accuracy is slightly lower. Fig.~\ref{fig:ssl_fig}(b) shows the increase of segmentation performance for RotNet with our method, as we visualize deeper layers of SSL Alexnet, using a supervised linear post-training as proposed by~\citet{asano2019critical}. This finding is aligned with the classification results by a single linear layer of~\citet{asano2019critical}, which improve with the layer index.

Tab.~\ref{tab:ssl_tab} compares RAP (which seems to be the strongest baseline) and our method in the predicted SSL class (not ImageNet) negative perturbation setting. Evidently, our method has superior performance. SSL results also seem correlated with the fully supervised ones (note that the architecture and the processing of the fully connected layer is different from the one used in Tab.~\ref{tab:neg_per_auc}). Two alternatives to our novel SSL procedure are also presented: in one, the difference from the neighbor is replaced with a sum. In the other, no comparison to the neighbor takes place. Our procedure for SSL explainability is superior for both our method and RAP.

\noindent\textbf{Ablation Study:}
By repeatedly employing normalization, our method is kept parameter-free. In Tab.~\ref{tab:ablation}, we present negative perturbation results for methods that are obtained by removing one components out of our complete method. We also present the results of a similar method in which the guided attribution based residual-term $\mathbf{r}$ is replaced by a GradCam term. As can be seen, each of these modifications damages the performance to some degree. Without any residual term, the method is slightly worse than RAP, while a partial residual term further hurts performance. In the supplementary material, we show visual results of the different components and variations of our method and present observations on the contribution of each part to the final outcome.

\section{Conclusions}
Explainability plays a major rule in debugging neural networks, creating trust in the predictive capabilities of networks beyond the specific test dataset, and in seeding downstream methods that analyze the images spatially. Previous visualization methods are either class-agnostic, low-resolution, or neglect much of the object's region, focusing on edges. The separation between relevant and irrelevant image parts provided by the previous methods is also often blurry. In this work, we present a novel explainability method that outputs class-dependent explanations that are clearer and more exact than those presented by the many existing methods tested. The new method is based on combining concepts from the two major branches of the current literature: attribution methods and gradient methods. This combination is done, on equal grounds, through the usage of a non-negative matrix factorization technique that partitions the image into foreground and background regions. Finally, we propose a novel procedure for evaluating the explainability of SSL methods.

\section*{Acknowledgment}
This project has received funding from the European Research Council (ERC) under the European Unions Horizon 2020 research and innovation programme (grant ERC CoG 725974). The contribution of the first author is part of a Ph.D. thesis research conducted at Tel Aviv University.

\bibliography{explainability}

\begin{thebibliography}{36}
\providecommand{\natexlab}[1]{#1}
\providecommand{\url}[1]{\texttt{#1}}
\providecommand{\urlprefix}{URL }
\expandafter\ifx\csname urlstyle\endcsname\relax
  \providecommand{\doi}[1]{doi:\discretionary{}{}{}#1}\else
  \providecommand{\doi}{doi:\discretionary{}{}{}\begingroup
  \urlstyle{rm}\Url}\fi

\bibitem[{Adebayo et~al.(2018)Adebayo, Gilmer, Muelly, Goodfellow, Hardt, and
  Kim}]{adebayo2018sanity}
Adebayo, J.; Gilmer, J.; Muelly, M.; Goodfellow, I.; Hardt, M.; and Kim, B.
  2018.
\newblock Sanity checks for saliency maps.
\newblock In \emph{Advances in Neural Information Processing Systems},
  9505--9515.

\bibitem[{Ahn, Cho, and Kwak(2019)}]{ahn2019weakly}
Ahn, J.; Cho, S.; and Kwak, S. 2019.
\newblock Weakly supervised learning of instance segmentation with inter-pixel
  relations.
\newblock In \emph{Proceedings of the IEEE Conference on Computer Vision and
  Pattern Recognition}, 2209--2218.

\bibitem[{Asano, Rupprecht, and
  Vedaldi(2019{\natexlab{a}})}]{asano2019critical}
Asano, Y.~M.; Rupprecht, C.; and Vedaldi, A. 2019{\natexlab{a}}.
\newblock A critical analysis of self-supervision, or what we can learn from a
  single image.
\newblock \emph{arXiv preprint arXiv:1904.13132} .

\bibitem[{Asano, Rupprecht, and Vedaldi(2019{\natexlab{b}})}]{asano2019self}
Asano, Y.~M.; Rupprecht, C.; and Vedaldi, A. 2019{\natexlab{b}}.
\newblock Self-labelling via simultaneous clustering and representation
  learning.
\newblock \emph{arXiv preprint arXiv:1911.05371} .

\bibitem[{Bach et~al.(2015)Bach, Binder, Montavon, Klauschen, M{\"u}ller, and
  Samek}]{bach2015pixel}
Bach, S.; Binder, A.; Montavon, G.; Klauschen, F.; M{\"u}ller, K.-R.; and
  Samek, W. 2015.
\newblock On pixel-wise explanations for non-linear classifier decisions by
  layer-wise relevance propagation.
\newblock \emph{PloS one} 10(7): e0130140.

\bibitem[{Binder et~al.(2016)Binder, Montavon, Lapuschkin, M{\"u}ller, and
  Samek}]{binder2016layer}
Binder, A.; Montavon, G.; Lapuschkin, S.; M{\"u}ller, K.-R.; and Samek, W.
  2016.
\newblock Layer-wise relevance propagation for neural networks with local
  renormalization layers.
\newblock In \emph{International Conference on Artificial Neural Networks},
  63--71. Springer.

\bibitem[{Dabkowski and Gal(2017)}]{dabkowski2017real}
Dabkowski, P.; and Gal, Y. 2017.
\newblock Real time image saliency for black box classifiers.
\newblock In \emph{Advances in Neural Information Processing Systems},
  6970--6979.

\bibitem[{Erhan et~al.(2009)Erhan, Bengio, Courville, and
  Vincent}]{erhan2009visualizing}
Erhan, D.; Bengio, Y.; Courville, A.; and Vincent, P. 2009.
\newblock Visualizing higher-layer features of a deep network.
\newblock \emph{University of Montreal} 1341(3): 1.

\bibitem[{Fong, Patrick, and Vedaldi(2019)}]{fong2019understanding}
Fong, R.; Patrick, M.; and Vedaldi, A. 2019.
\newblock Understanding deep networks via extremal perturbations and smooth
  masks.
\newblock In \emph{Proceedings of the IEEE International Conference on Computer
  Vision}, 2950--2958.

\bibitem[{Fong and Vedaldi(2017)}]{fong2017interpretable}
Fong, R.~C.; and Vedaldi, A. 2017.
\newblock Interpretable explanations of black boxes by meaningful perturbation.
\newblock In \emph{Proceedings of the IEEE International Conference on Computer
  Vision}, 3429--3437.

\bibitem[{Gao et~al.(2016)Gao, Chen, Zheng, and Fang}]{gao2016factorization}
Gao, M.; Chen, H.; Zheng, S.; and Fang, B. 2016.
\newblock A factorization based active contour model for texture segmentation.
\newblock In \emph{2016 IEEE International Conference on Image Processing
  (ICIP)}, 4309--4313. IEEE.

\bibitem[{Gu, Yang, and Tresp(2018)}]{gu2018understanding}
Gu, J.; Yang, Y.; and Tresp, V. 2018.
\newblock Understanding individual decisions of cnns via contrastive
  backpropagation.
\newblock In \emph{Asian Conference on Computer Vision}, 119--134. Springer.

\bibitem[{Guillaumin, K{\"u}ttel, and Ferrari(2014)}]{guillaumin2014ImageNet}
Guillaumin, M.; K{\"u}ttel, D.; and Ferrari, V. 2014.
\newblock Imagenet auto-annotation with segmentation propagation.
\newblock \emph{International Journal of Computer Vision} 110(3): 328--348.

\bibitem[{Hoyer et~al.(2019)Hoyer, Munoz, Katiyar, Khoreva, and
  Fischer}]{hoyer2019grid}
Hoyer, L.; Munoz, M.; Katiyar, P.; Khoreva, A.; and Fischer, V. 2019.
\newblock Grid saliency for context explanations of semantic segmentation.
\newblock In \emph{Advances in Neural Information Processing Systems},
  6462--6473.

\bibitem[{Huang et~al.(2018)Huang, Wang, Wang, Liu, and Wang}]{huang2018weakly}
Huang, Z.; Wang, X.; Wang, J.; Liu, W.; and Wang, J. 2018.
\newblock Weakly-supervised semantic segmentation network with deep seeded
  region growing.
\newblock In \emph{Proceedings of the IEEE Conference on Computer Vision and
  Pattern Recognition}, 7014--7023.

\bibitem[{Iwana, Kuroki, and Uchida(2019)}]{iwana2019explaining}
Iwana, B.~K.; Kuroki, R.; and Uchida, S. 2019.
\newblock Explaining Convolutional Neural Networks using Softmax Gradient
  Layer-wise Relevance Propagation.
\newblock \emph{arXiv preprint arXiv:1908.04351} .

\bibitem[{Kindermans et~al.(2017)Kindermans, Sch{\"u}tt, Alber, M{\"u}ller,
  Erhan, Kim, and D{\"a}hne}]{kindermans2017learning}
Kindermans, P.-J.; Sch{\"u}tt, K.~T.; Alber, M.; M{\"u}ller, K.-R.; Erhan, D.;
  Kim, B.; and D{\"a}hne, S. 2017.
\newblock Learning how to explain neural networks: Patternnet and
  patternattribution.
\newblock \emph{arXiv preprint arXiv:1705.05598} .

\bibitem[{Lundberg and Lee(2017)}]{lundberg2017unified}
Lundberg, S.~M.; and Lee, S.-I. 2017.
\newblock A unified approach to interpreting model predictions.
\newblock In \emph{Advances in Neural Information Processing Systems},
  4765--4774.

\bibitem[{Mahendran and Vedaldi(2016)}]{mahendran2016visualizing}
Mahendran, A.; and Vedaldi, A. 2016.
\newblock Visualizing deep convolutional neural networks using natural
  pre-images.
\newblock \emph{International Journal of Computer Vision} 120(3): 233--255.

\bibitem[{Montavon et~al.(2017)Montavon, Lapuschkin, Binder, Samek, and
  M{\"u}ller}]{montavon2017explaining}
Montavon, G.; Lapuschkin, S.; Binder, A.; Samek, W.; and M{\"u}ller, K.-R.
  2017.
\newblock Explaining nonlinear classification decisions with deep taylor
  decomposition.
\newblock \emph{Pattern Recognition} 65: 211--222.

\bibitem[{Nam et~al.(2019)Nam, Gur, Choi, Wolf, and Lee}]{nam2019relative}
Nam, W.-J.; Gur, S.; Choi, J.; Wolf, L.; and Lee, S.-W. 2019.
\newblock Relative Attributing Propagation: Interpreting the Comparative
  Contributions of Individual Units in Deep Neural Networks.
\newblock \emph{arXiv preprint arXiv:1904.00605} .

\bibitem[{Russakovsky et~al.(2015)Russakovsky, Deng, Su, Krause, Satheesh, Ma,
  Huang, Karpathy, Khosla, Bernstein et~al.}]{russakovsky2015ImageNet}
Russakovsky, O.; Deng, J.; Su, H.; Krause, J.; Satheesh, S.; Ma, S.; Huang, Z.;
  Karpathy, A.; Khosla, A.; Bernstein, M.; et~al. 2015.
\newblock Imagenet large scale visual recognition challenge.
\newblock \emph{International journal of computer vision} 115(3): 211--252.

\bibitem[{Selvaraju et~al.(2017)Selvaraju, Cogswell, Das, Vedantam, Parikh, and
  Batra}]{selvaraju2017grad}
Selvaraju, R.~R.; Cogswell, M.; Das, A.; Vedantam, R.; Parikh, D.; and Batra,
  D. 2017.
\newblock Grad-cam: Visual explanations from deep networks via gradient-based
  localization.
\newblock In \emph{Proceedings of the IEEE international conference on computer
  vision}, 618--626.

\bibitem[{Shrikumar, Greenside, and Kundaje(2017)}]{shrikumar2017learning}
Shrikumar, A.; Greenside, P.; and Kundaje, A. 2017.
\newblock Learning important features through propagating activation
  differences.
\newblock In \emph{Proceedings of the 34th International Conference on Machine
  Learning-Volume 70}, 3145--3153. JMLR. org.

\bibitem[{Shrikumar et~al.(2016)Shrikumar, Greenside, Shcherbina, and
  Kundaje}]{shrikumar2016not}
Shrikumar, A.; Greenside, P.; Shcherbina, A.; and Kundaje, A. 2016.
\newblock Not just a black box: Learning important features through propagating
  activation differences.
\newblock \emph{arXiv preprint arXiv:1605.01713} .

\bibitem[{Simonyan, Vedaldi, and Zisserman(2013)}]{simonyan2013deep}
Simonyan, K.; Vedaldi, A.; and Zisserman, A. 2013.
\newblock Deep inside convolutional networks: Visualising image classification
  models and saliency maps.
\newblock \emph{arXiv preprint arXiv:1312.6034} .

\bibitem[{Smilkov et~al.(2017)Smilkov, Thorat, Kim, Vi{\'e}gas, and
  Wattenberg}]{smilkov2017smoothgrad}
Smilkov, D.; Thorat, N.; Kim, B.; Vi{\'e}gas, F.; and Wattenberg, M. 2017.
\newblock Smoothgrad: removing noise by adding noise.
\newblock \emph{arXiv preprint arXiv:1706.03825} .

\bibitem[{Srinivas and Fleuret(2019)}]{srinivas2019full}
Srinivas, S.; and Fleuret, F. 2019.
\newblock Full-gradient representation for neural network visualization.
\newblock In \emph{Advances in Neural Information Processing Systems},
  4126--4135.

\bibitem[{Sundararajan, Taly, and Yan(2017)}]{sundararajan2017axiomatic}
Sundararajan, M.; Taly, A.; and Yan, Q. 2017.
\newblock Axiomatic attribution for deep networks.
\newblock In \emph{Proceedings of the 34th International Conference on Machine
  Learning-Volume 70}, 3319--3328. JMLR. org.

\bibitem[{Van~Gansbeke et~al.(2020)Van~Gansbeke, Vandenhende, Georgoulis,
  Proesmans, and Van~Gool}]{vanscan}
Van~Gansbeke, W.; Vandenhende, S.; Georgoulis, S.; Proesmans, M.; and Van~Gool,
  L. 2020.
\newblock SCAN: Learning to Classify Images without Labels.
\newblock In \emph{European Conference on Computer Vision (ECCV)}.

\bibitem[{Wang et~al.(2019)Wang, Zhang, Kan, Shan, and Chen}]{wang2019self}
Wang, Y.; Zhang, J.; Kan, M.; Shan, S.; and Chen, X. 2019.
\newblock Self-supervised Scale Equivariant Network for Weakly Supervised
  Semantic Segmentation.
\newblock \emph{arXiv preprint arXiv:1909.03714} .

\bibitem[{Yuan, Wang, and Cheriyadat(2015)}]{yuan2015factorization}
Yuan, J.; Wang, D.; and Cheriyadat, A.~M. 2015.
\newblock Factorization-based texture segmentation.
\newblock \emph{IEEE Transactions on Image Processing} 24(11): 3488--3497.

\bibitem[{Zeiler and Fergus(2014)}]{zeiler2014visualizing}
Zeiler, M.~D.; and Fergus, R. 2014.
\newblock Visualizing and understanding convolutional networks.
\newblock In \emph{European conference on computer vision}, 818--833. Springer.

\bibitem[{Zhang et~al.(2018)Zhang, Bargal, Lin, Brandt, Shen, and
  Sclaroff}]{zhang2018top}
Zhang, J.; Bargal, S.~A.; Lin, Z.; Brandt, J.; Shen, X.; and Sclaroff, S. 2018.
\newblock Top-down neural attention by excitation backprop.
\newblock \emph{International Journal of Computer Vision} 126(10): 1084--1102.

\bibitem[{Zhou et~al.(2018)Zhou, Bau, Oliva, and
  Torralba}]{zhou2018interpreting}
Zhou, B.; Bau, D.; Oliva, A.; and Torralba, A. 2018.
\newblock Interpreting deep visual representations via network dissection.
\newblock \emph{IEEE transactions on pattern analysis and machine intelligence}
  .

\bibitem[{Zhou et~al.(2016)Zhou, Khosla, Lapedriza, Oliva, and
  Torralba}]{zhou2016learning}
Zhou, B.; Khosla, A.; Lapedriza, A.; Oliva, A.; and Torralba, A. 2016.
\newblock Learning deep features for discriminative localization.
\newblock In \emph{Proceedings of the IEEE conference on computer vision and
  pattern recognition}, 2921--2929.

\end{thebibliography}

\clearpage
\onecolumn
\appendix 
\title{Supplementary: Differentiable Gaussian PSF Layer}
\author{}
\maketitle
\setcounter{equation}{0}
\setcounter{figure}{0}
\section{Code}
\begin{figure*}[ht]
    \centering
    \includegraphics[width=0.73\linewidth]{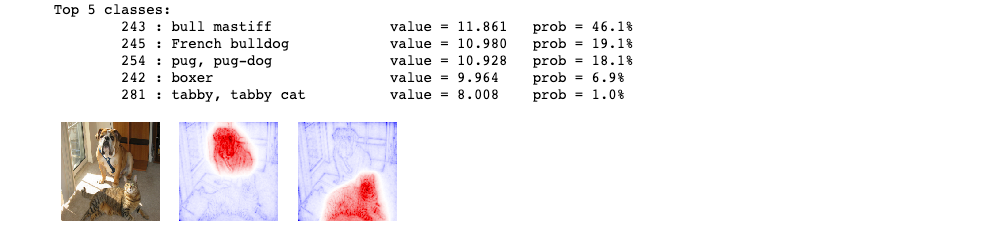}
    \caption{Visualization of two different classes for VGG-19.}
    \label{fig:jup_multi}
    \centering
    \includegraphics[width=0.73\linewidth]{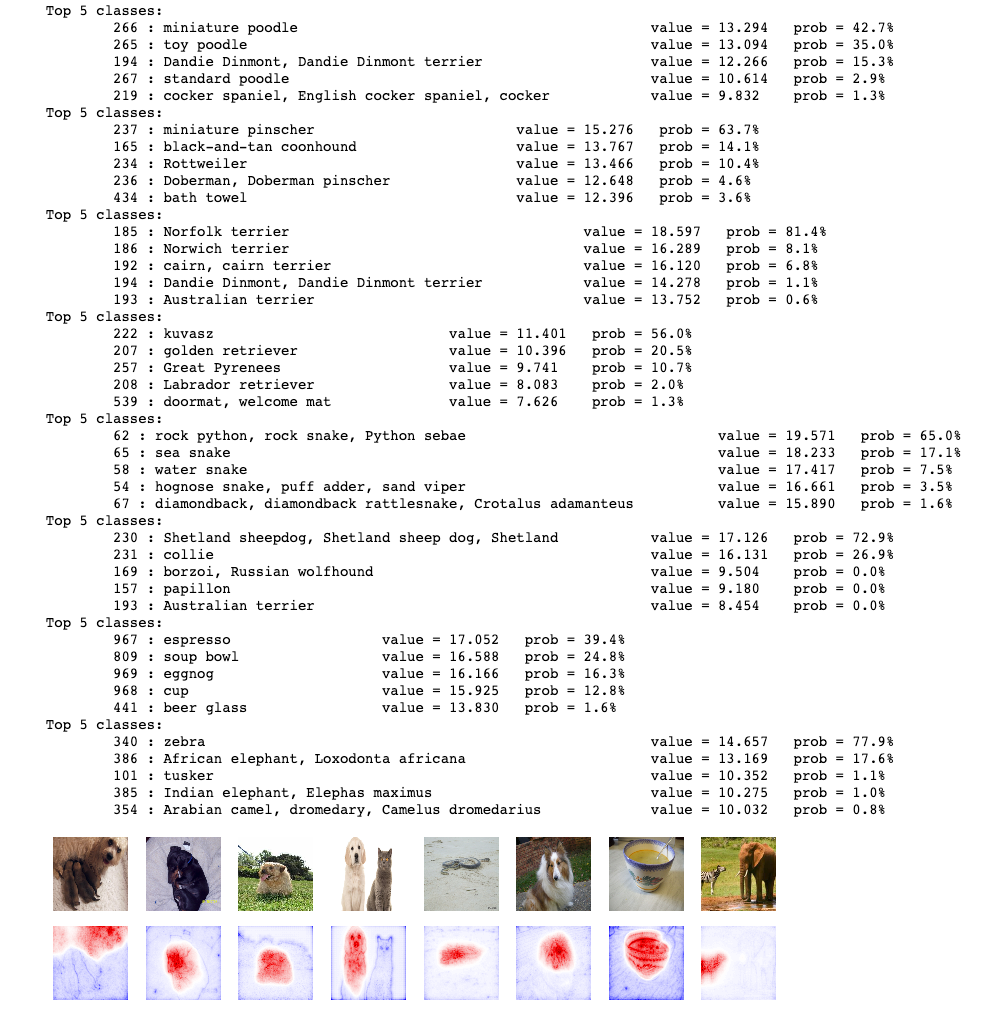}
    \caption{Visualization of the top predicted class of a VGG-19 ImageNet trained network.}
    \label{fig:jup_top}
\end{figure*}

The code is written in PyTorch, and is provided as part of the supplementary material, including a {\em Jupyter} notebook with examples. Fig.~\ref{fig:jup_multi},\ref{fig:jup_top} show screenshots from the notebook.

\clearpage
\section{Algorithm}
Our method is listed in Alg.~\ref{alg:algorithm}.\\
Note that for the linear layers of the network, we compute the residual using only $\mathbf{F}_{x\nabla x}$ in the following way:
\begin{align}
    \mathbf{r}^{(n)} &= 
    \begin{cases}
        \mathbf{M}^{(n)}_{x\nabla x}, &\text{First linear layer}\\
        0, &\text{Otherwise}
    \end{cases}
\end{align}
We note that $x$ and $\nabla x$ are reshaped into their previous (2D) form.
\begin{algorithm}
\caption{Class Attribution Propagation with Guided Factorization.}\label{alg:algorithm}
\begin{algorithmic}[1]
\Require $M:$ Neural network model, $I\in\mathbb{R}^{C \times H \times W}:$ Input images, $t\in\mathcal{C}:$ Target class.
\State $M(I)$ \Comment{Forward-pass, save intermediate feature maps}
\State $\sigma \leftarrow \max\|y-y^t\|_1$
\State $\hat{y} \leftarrow  \operatorname{softmax} \big(y^t \exp{\big(- \frac{1}{2}\big(\frac{y-y^t}{\sigma}\big)^2\big)}\big)$
\State $\Phi^{(1)} \leftarrow  x^{(1)} \odot \frac{\partial \hat{y}^t}{\partial x^{(1)}}$ \Comment{Initial Attribution - First Linear layer}

\For{linear layers,  $n > 1$}
\State $\mathbf{C}_j^{(n)} \leftarrow \mathcal{G}^{(n)}(|x^{(n)}|, |\theta^{(n)}|, \Phi^{(n-1)})$ \Comment{Absolute influence }

\If{next layer is 2D}
\State Reshape $x$ and $\nabla x$ to the previous 2D form
\State $\mathbf{M}^{(n)} \leftarrow  \mathcal{N}_{\max}\bigg(\bigg(\frac{1}{|C|}\sum_{c \in [C]}\big(x_c^{(n)} \odot \nabla x_c^{(n)}\big)\bigg)^+\bigg)$
\State $\mathbf{r}^{(n)} \leftarrow  \mathbf{M}^{(n)}$ \Comment{Residual}
\Else{}
\State $\mathbf{r}^{(n)} \leftarrow 0$
\EndIf

\State $\Phi^{(n)} \leftarrow  \Delta^{(n)}_\text{\em shift}(\mathbf{C}^{(n)}, \mathbf{r}^{(n)})$ \Comment{Shifting by the residual}
\EndFor

\For{convolution layers}
\State Compute $\mathbf{C}^{(n)}$, $\mathbf{M}^{(n)}$
\State $\mathbf{A}^{(n)}_j \leftarrow \mathcal{G}^{(n)}(\mathbb{1}, |\theta^{(n)}|, \Phi^{(n-1)})$ \Comment{Input agnostic attribution}
\State $\mathbf{F}^{(n)}_{x} \leftarrow \bar{\mathcal{F}}(x^{(n)},\mathbf{C}^{(n)})^+$ \Comment{Factorization of input feature map}
\State $\mathbf{F}^{(n)}_{\nabla x} \leftarrow \bar{\mathcal{F}}(\nabla x^{(n)},\mathbf{C}^{(n)})^+$ \Comment{Factorization of input feature map gradients}
\State $\mathbf{r}^{(n)} \leftarrow \mathbf{A}^{(n)} + \mathbf{F}^{(n)}_{\nabla x} + \frac{\mathbf{F}^{(n)}_{x} + \mathbf{M}^{(n)}_{x\nabla x}}{1 + \exp(-\mathbf{C}^{(n)})}$ \Comment{Residual}
\State Compute $\Phi^{(n)}$
\EndFor
\end{algorithmic}
\end{algorithm}

\clearpage
\section{Ablation Study}
\begin{figure}[h]
    \centering
    \includegraphics[width=\linewidth]{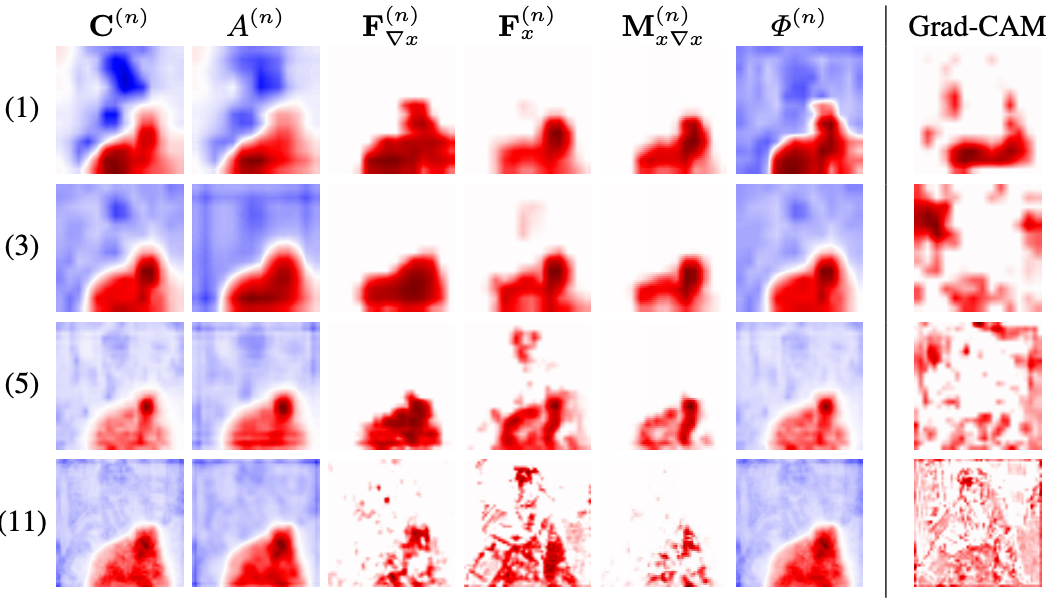}
    \caption{Method decomposition on the image of Fig.1(a) of the paper. Visualization of the different elements of the method for a pre-trained VGG-19 network. Predicted top-class is ``dog'', propagated class is ``cat''. (1, 5, 11) - Convolution layer number.}
    \label{fig:decomp}
\end{figure}

By repeatedly employing normalization, our method is kept parameter-free. The different components of it, presented in Sec.4 of the paper, are visualized in Fig~\ref{fig:decomp}. One can observe several behaviors of the different component, (i) across all components, the semantic information is best visible in deeper layers of the network, where residual information is becoming more texture-oriented at shallower layers. (ii) The difference between $F_x$ and $F_{\nabla x}$ is mostly visible in the out-of-class regions: $F_x$ is derived from the data directly and is biased toward input-specific activations, resulting in highlights from the class that is not being visualized from layer 5 onward. (iii) The input-agnostic data term $\mathbf{A}$ is more blurry than $\mathbf{C}$, as a result of using the $\mathbb{1}$ tensor as input. It can, therefore, serve as a regularization term that is less susceptible to image edges. 

\clearpage
\section{Additional Results}
In Tab.~\ref{tab:neg_per_auc_sup} and~\ref{tab:segmentation_sup} we show additional results for two different Grad-CAM variants, considering visualisation from shallower layers as denoted by {Grad-CAM* = 4-layers shallower and Grad-CAM** = 8-layers shallower. An example of the different outputs can be seen in Fig.~\ref{fig:decomp} of the previous section, where the semantic information of the Grad=CAM is less visible as we go shallower, where as in our method, the sematic information is becoming more fine grained.} Additionally, we show the inferior results of Gradient SHAP and DeepLIFT SHAP for negative perturbation.
\begin{table}[t]
    \caption{Area Under the Curve (AUC) values for the two negative perturbation tests. Grad-CAM variants : * = 4-layers shallower, ** = 8-layer shallower.}
    \begin{tabular*}{\textwidth}{@{\extracolsep{\fill}}lcccccc}
        \toprule
         & Gradient & DeepLIFT  & \multirow{2}{*}{Grad-CAM} & \multirow{2}{*}{Grad-CAM*} & \multirow{2}{*}{Grad-CAM**} & \multirow{2}{*}{Ours} \\
         & SHAP & SHAP & & & &\\
         \midrule
        Predicted & 7.5 & 7.8 & 37.8 & 29.9 & 16.0 & \textbf{38.9}\\
        Target    & 7.7 & 8.1 & 39.4 & 30.9 & 16.5 & \textbf{40.0}\\
        \bottomrule
    \end{tabular*}
\label{tab:neg_per_auc_sup}
\end{table}
\begin{table}[t]
    \caption{Quantitative segmentation results on (a) ImageNet and (b) PASCAL-VOC 2012.}
    \begin{tabular*}{\textwidth}{@{\extracolsep{\fill}}c@{~~~}l@{~~~}c@{~~~}c@{~~~}c@{~~~}c@{}}
        \toprule
         & & Grad-CAM & Grad-CAM* & Grad-CAM** & Ours \\
        \midrule
        \multirow{2}{*}{ImageNet} &Pixel Acc. & 66.5 & 65.5 & 50.5 & \textbf{79.9}\\
        & mAP         & 62.4 & 61.6 & 51.1 & \textbf{76.7}\\
        
        \midrule
        
        \multirow{2}{*}{VOC'12} &Pixel Acc. & 52.6 & 18.4 & 17.2 & \textbf{77.5}\\
        & mAP        & 41.7 & 44.5 & 43.4 & \textbf{45.9}\\
        \bottomrule
    \end{tabular*}
    \vspace{-.5cm}
    \label{tab:segmentation_sup}
\end{table}

\clearpage
\section{Self-Supervised Learning}
Our submission is the first contribution, as far as we can ascertain, that employs explainability methods on Self Supervised Learning methods. We evaluate our method on three recent state-of-the-art models. First, we show our method performance on AlexNet model, trained by RotNet in a completely self-supervised fashion which depends only on data augmentations specially predicting the 2d image rotations that is applied to the image that it gets as input. In order to evaluate the quality of the features learned we follow (only for the RotNet experiments) a similar approach used in the literature by using a supervised linear post-training in order to visualize deeper layers. 

\begin{figure}[H]
    \centering
    \subfigure[]{\includegraphics[width=0.16\textwidth]{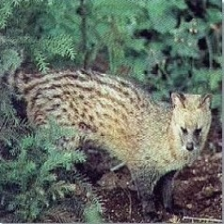}} 
    \subfigure[]{\includegraphics[width=0.16\textwidth]{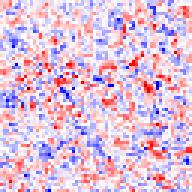}} 
    \subfigure[]{\includegraphics[width=0.16\textwidth]{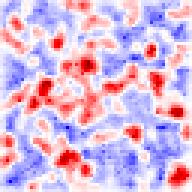}}
    \subfigure[]{\includegraphics[width=0.16\textwidth]{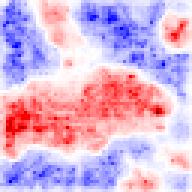}}
    \subfigure[]{\includegraphics[width=0.16\textwidth]{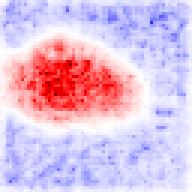}}
    \subfigure[]{\includegraphics[width=0.16\textwidth]{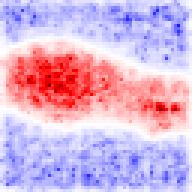}}
    \caption{The visualization after each layer. (a) original image. (b) visualization after first layer. (c) visualization after second layer. (d) visualization after third layer. (e) visualization after the fourth layer. (e) visualization after last layer. }
    \label{fig:perlayer}
\end{figure}

Fig.~\ref{fig:perlayer} shows the visualization obtained for the image on the left when training a supervised linear classifier after each layer and applying our method. As can be seen, early layers learns edges-like patterns and the deeper layer learns more complicated patterns, more visualizations are shown in the end of the supplementary.

In order to evaluate SSL methods without any additional supervision, we employ the classifier the SSL is trained with. This is presented in Sec.~4.4 of the paper and is given below as pseudocode in Alg~\ref{alg:ssl}.
\begin{algorithm}[H]
 \caption{self-supervised explainability by adopting nearest neighbors}
 \label{alg:ssl}
 \begin{algorithmic}[1]
 \Require $I$ input image, $\mathcal{S}$ set of images, $M=\phi_{\tau} \circ \phi_F$ the SSL network, where $\phi_F$ extracts the features and $\phi_{\tau}$ is the linear SSL classifier.
  \State $L_I = \phi_F(I)$
 \Comment{$L_I$ is the latent vector of image $I$}
 \State $\mathcal{L} \leftarrow \{ \phi_F(J) | J\in \mathcal{S} \}$
 \Comment{$\mathcal{L}$ is the set of all latent vectors for all images in $\mathcal{S}$}
 \State $L_N = \arg\min_{L\in\mathcal{L}} \| L_{I} -L\|$
 \Comment{$L_N$ is the nearest neighbor of $L$}
 \State ${S}_I = L_I - L_N$
 \Comment{Subtracting $L_N$ from $L_I$ to emphasize the unique elements of $L_I$}
 \State $v = \phi_{\tau}({S}_I)$
 \Comment{Forward pass with the new latent vector}
 \State $t = \arg\max v$
 \Comment{Choose the class with the highest probability}
 \State Apply Alg.~\ref{alg:algorithm}, with the input tuple $[\phi_{\tau}(S_I) , \phi_F(I)] , I , t$ 
  \end{algorithmic}
\end{algorithm}

\clearpage
\section{Additional Results - Multi Label}
\begin{figure}[h]
    \centering
    \includegraphics[width=\textwidth]{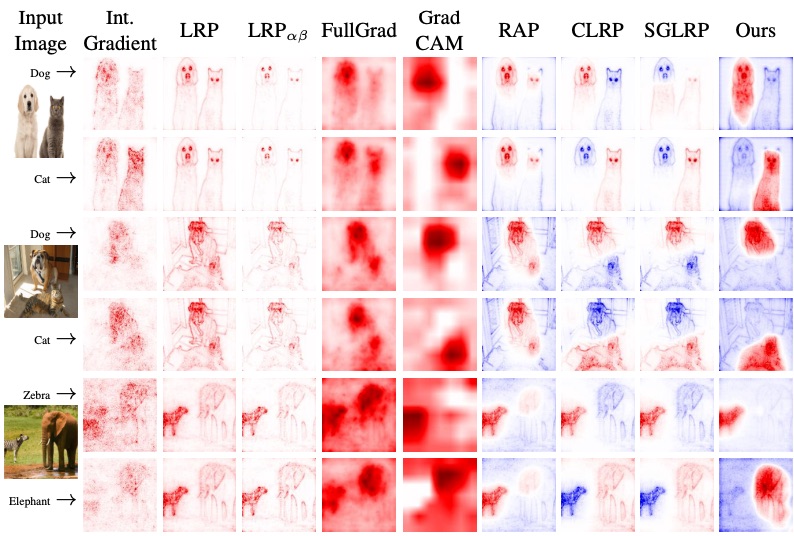}
    \caption{Visualization of two different classes for VGG-19.}
    \label{fig:multi_class_1}
\end{figure}
\begin{figure}
    \centering
    \includegraphics[width=\textwidth]{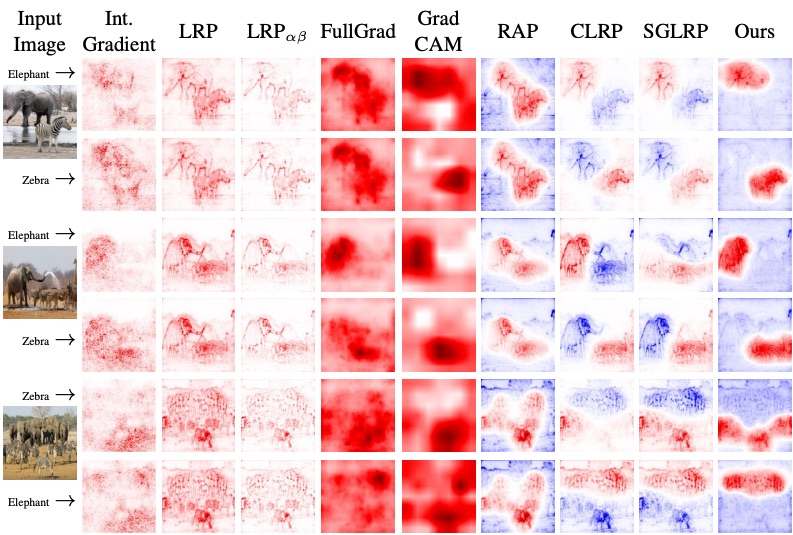}
    \caption{Visualization of two different classes for VGG-19.}
    \label{fig:multi_class_1}
\end{figure}

\clearpage
\section{Additional Results - Top Class}
\begin{figure}[h]
    \centering
    \includegraphics[width=\textwidth]{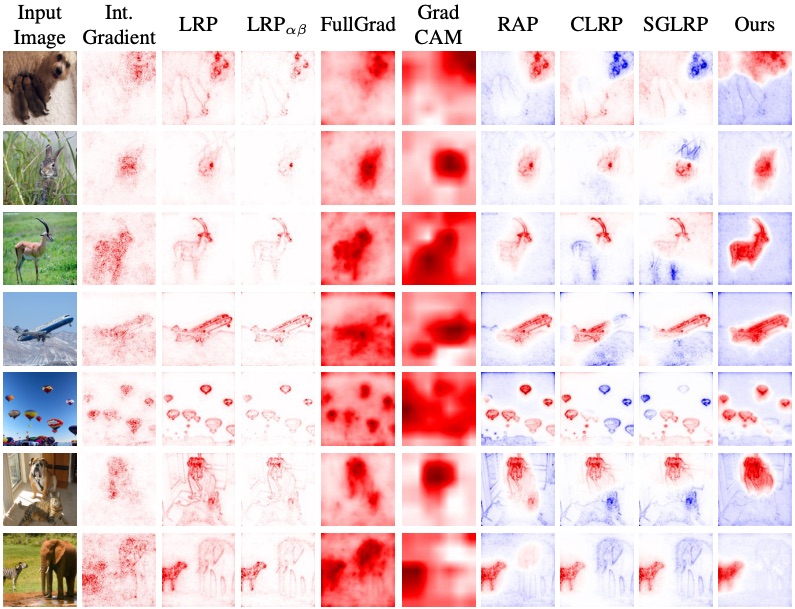}
    \caption{Visualization of the top predicted class of a VGG-19 ImageNet trained network.}
    \label{fig:top_class_1}
\end{figure}

\clearpage
\begin{figure}
    \centering
    \includegraphics[width=\textwidth]{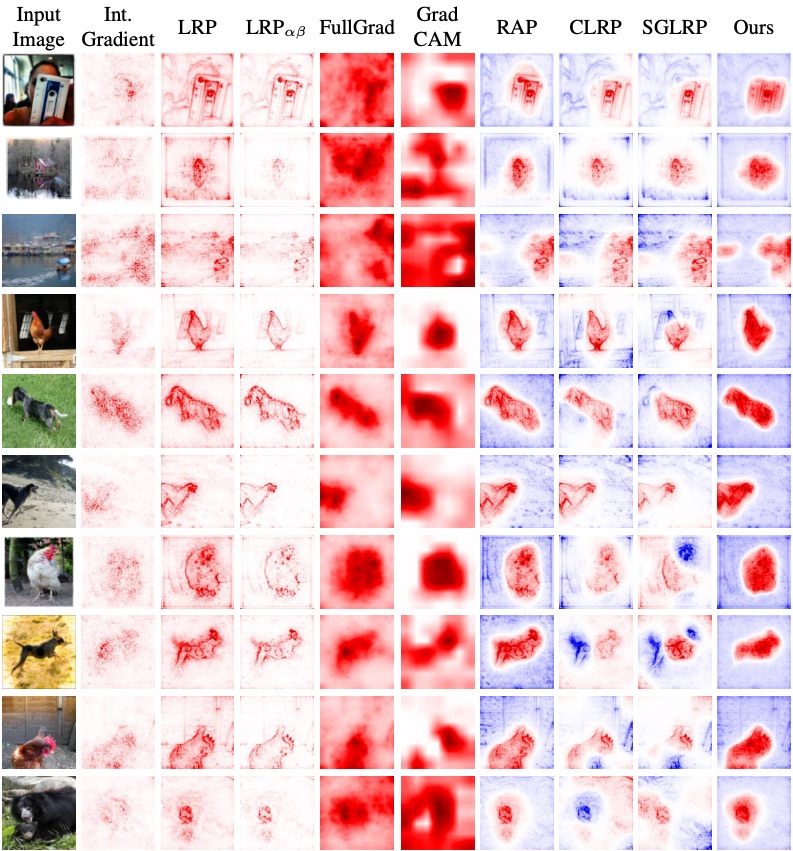}
    \caption{Visualization of the top predicted class of a VGG-19 ImageNet trained network.}
    \label{fig:top_class_2}
\end{figure}
\clearpage
\begin{figure}
    \centering
    \includegraphics[width=\textwidth]{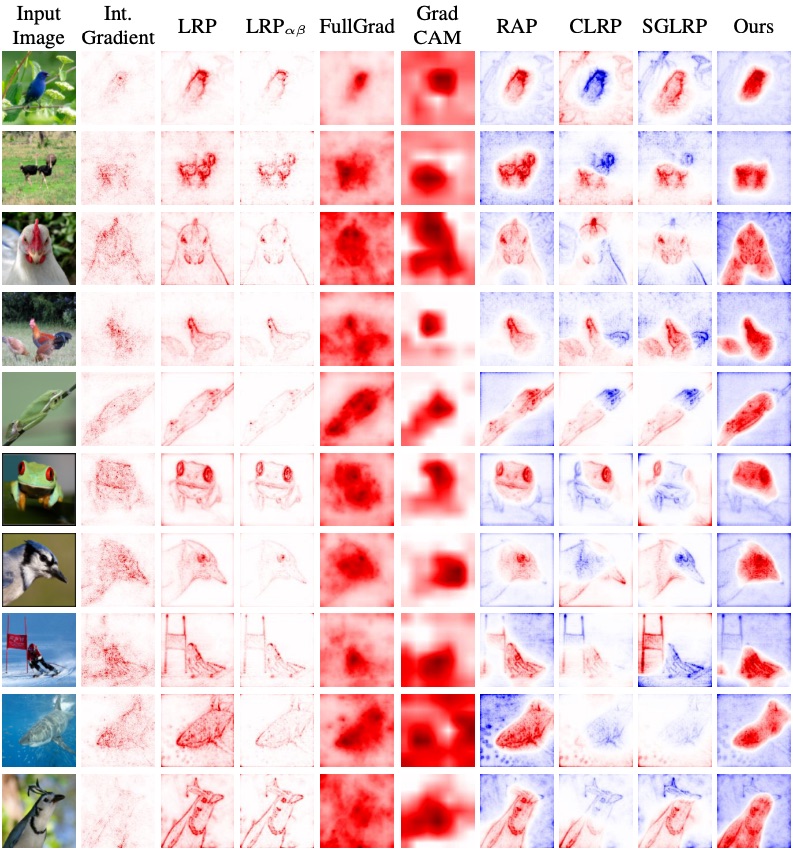}
    \caption{Visualization of the top predicted class of a VGG-19 ImageNet trained network.}
    \label{fig:top_class_3}
\end{figure}
\clearpage
\begin{figure}
    \centering
    \includegraphics[width=\textwidth]{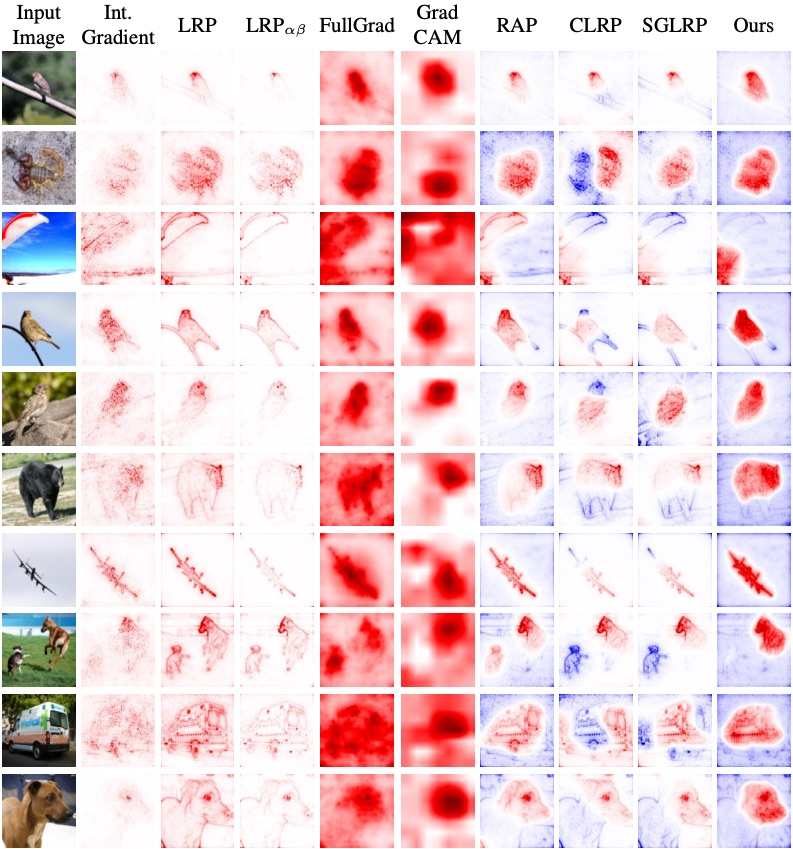}
    \caption{Visualization of the top predicted class of a VGG-19 ImageNet trained network.}
    \label{fig:top_class_4}
\end{figure}
\clearpage
\begin{figure}
    \centering
    \includegraphics[width=\textwidth]{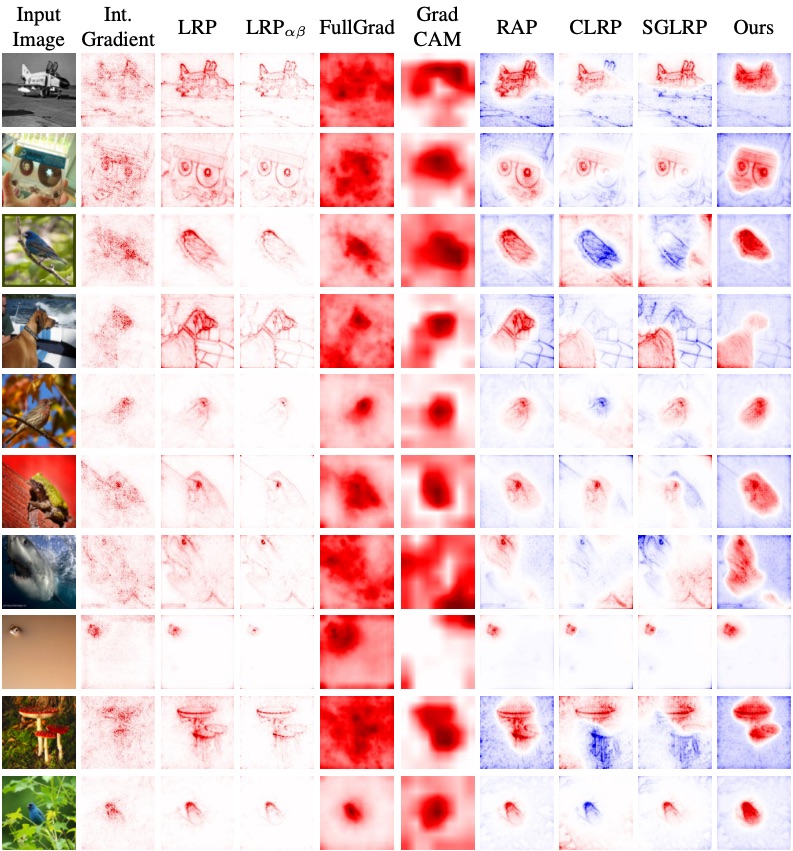}
    \caption{Visualization of the top predicted class of a VGG-19 ImageNet trained network.}
    \label{fig:top_class_5}
\end{figure}
\clearpage
\begin{figure}
    \centering
    \includegraphics[width=\textwidth]{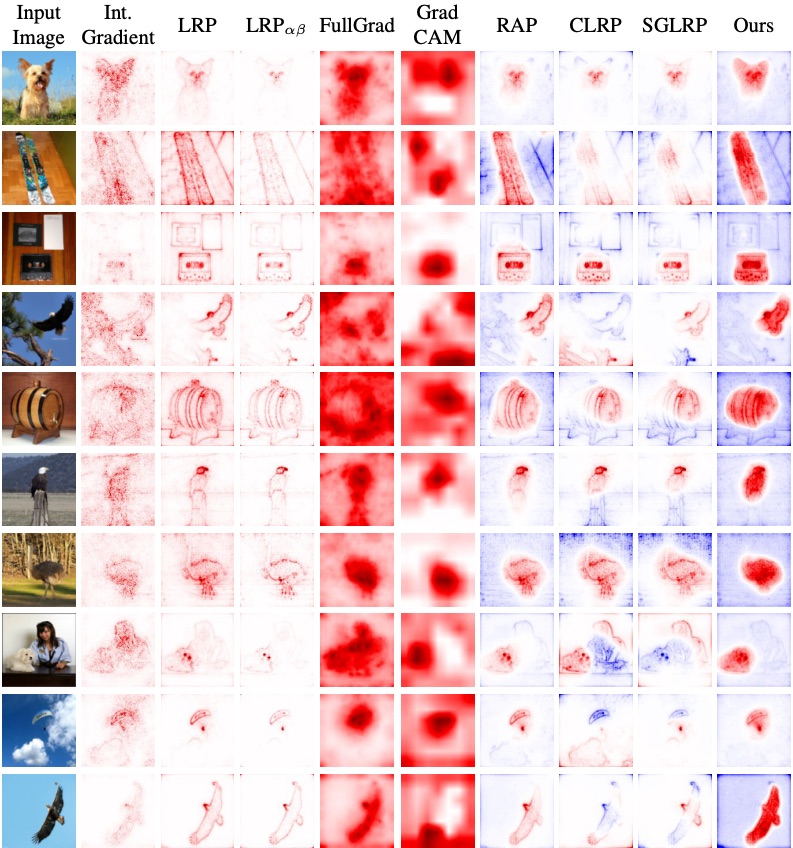}
    \caption{Visualization of the top predicted class of a VGG-19 ImageNet trained network.}
    \label{fig:top_class_6}
\end{figure}
\clearpage
\begin{figure}
    \centering
    \includegraphics[width=\textwidth]{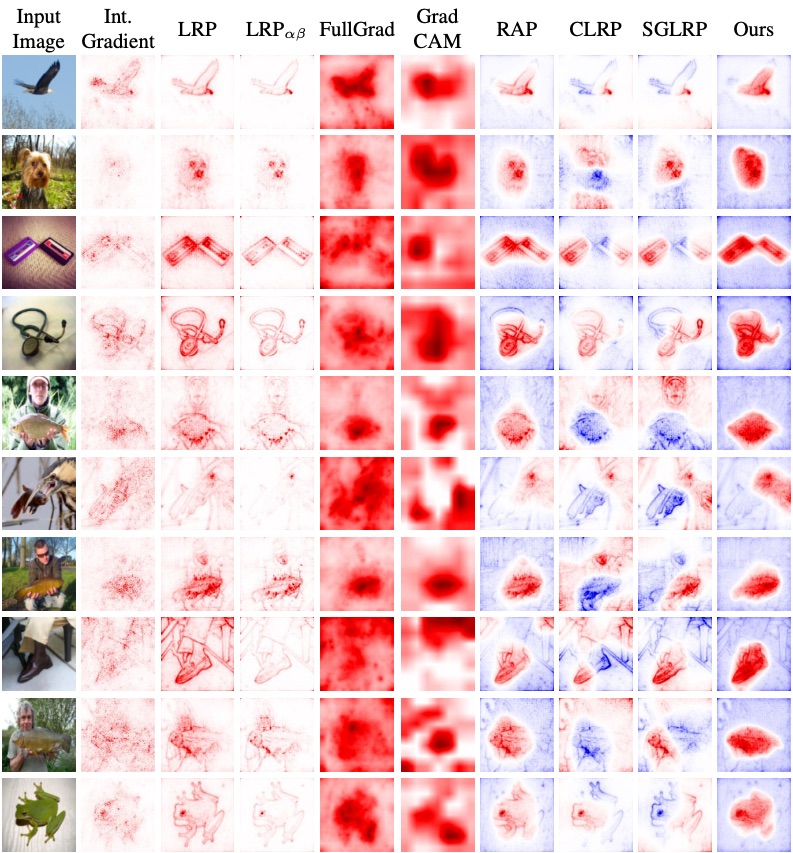}
    \caption{Visualization of the top predicted class of a VGG-19 ImageNet trained network.}
    \label{fig:top_class_7}
\end{figure}
\clearpage
\begin{figure}
    \centering
    \includegraphics[width=\textwidth]{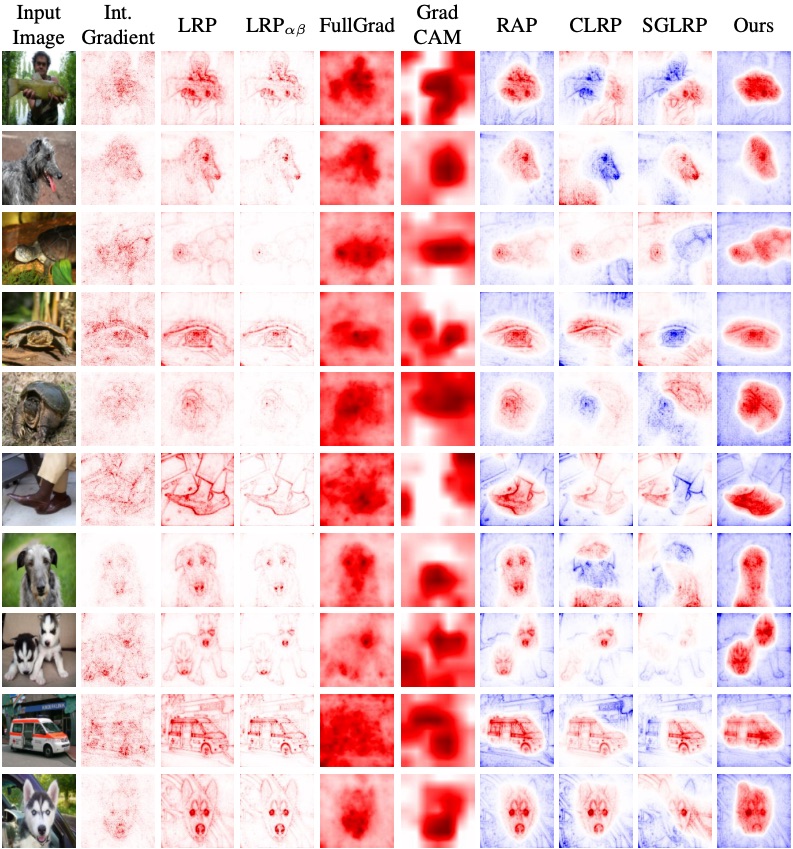}
    \caption{Visualization of the top predicted class of a VGG-19 ImageNet trained network.}
    \label{fig:top_class_8}
\end{figure}

\clearpage
\begin{figure}
    \centering
    \includegraphics[width=\textwidth]{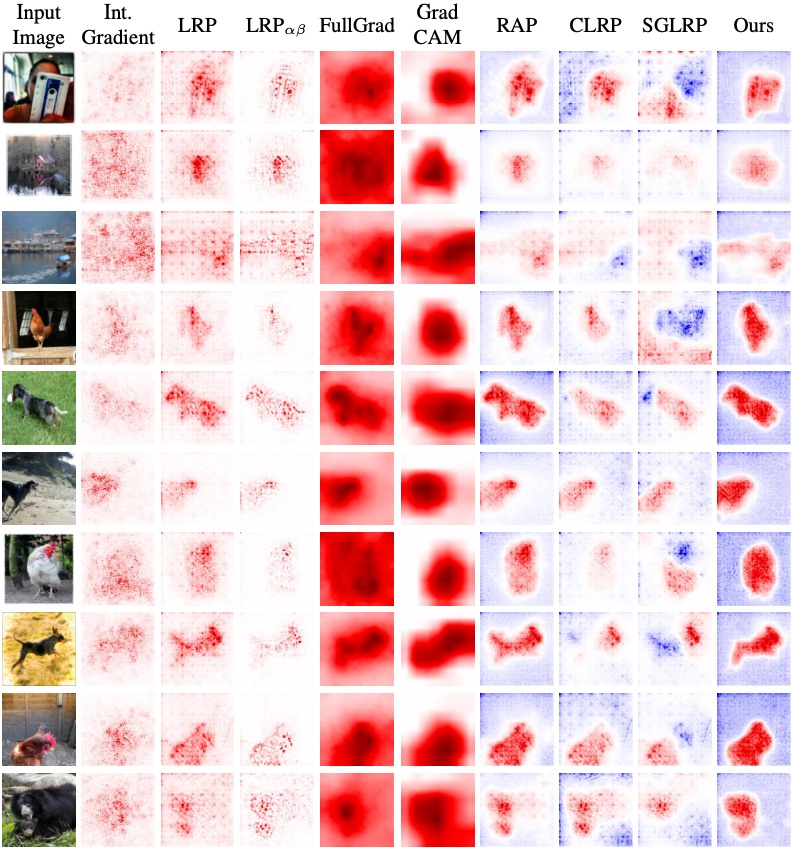}
    \caption{Visualization of the top predicted class of a ResNet-50 ImageNet trained network.}
    \label{fig:top_class_9}
\end{figure}

\clearpage
\begin{figure}
    \centering
    \includegraphics[width=\textwidth]{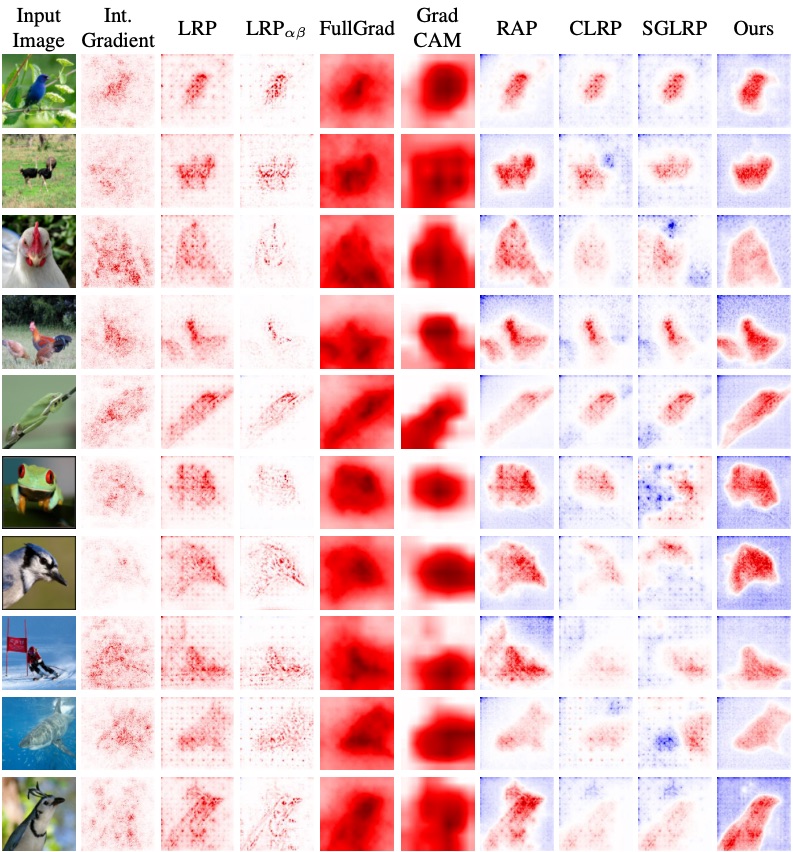}
    \caption{Visualization of the top predicted class of a ResNet-50 ImageNet trained network.}
    \label{fig:top_class_10}
\end{figure}
\clearpage
\begin{figure}
    \centering
    \includegraphics[width=\textwidth]{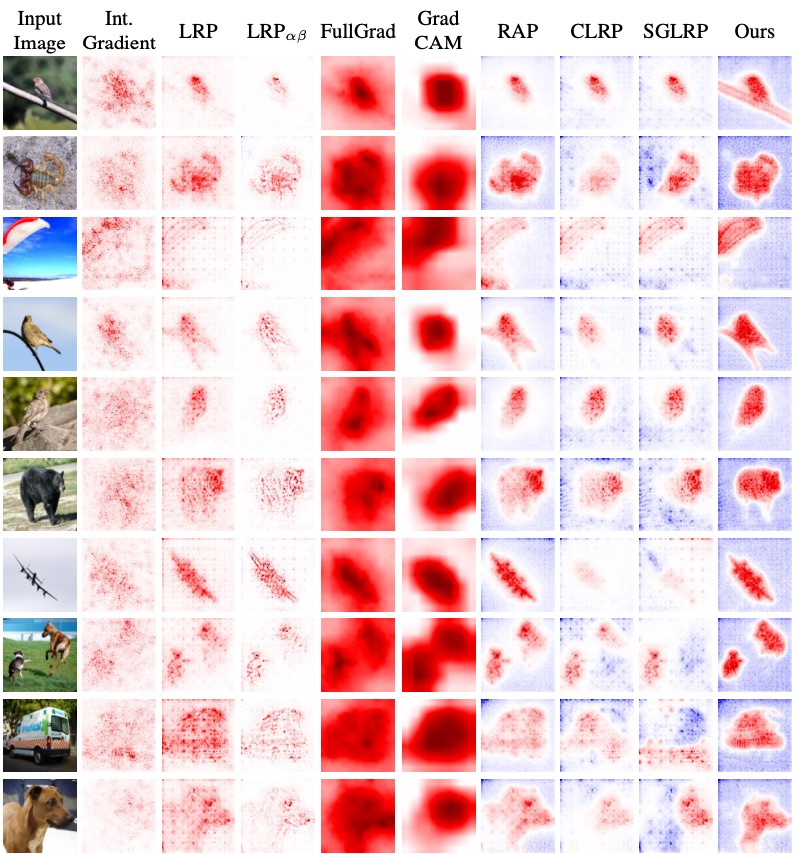}
    \caption{Visualization of the top predicted class of a ResNet-50 ImageNet trained network.}
    \label{fig:top_class_11}
\end{figure}
\clearpage
\begin{figure}
    \centering
    \includegraphics[width=\textwidth]{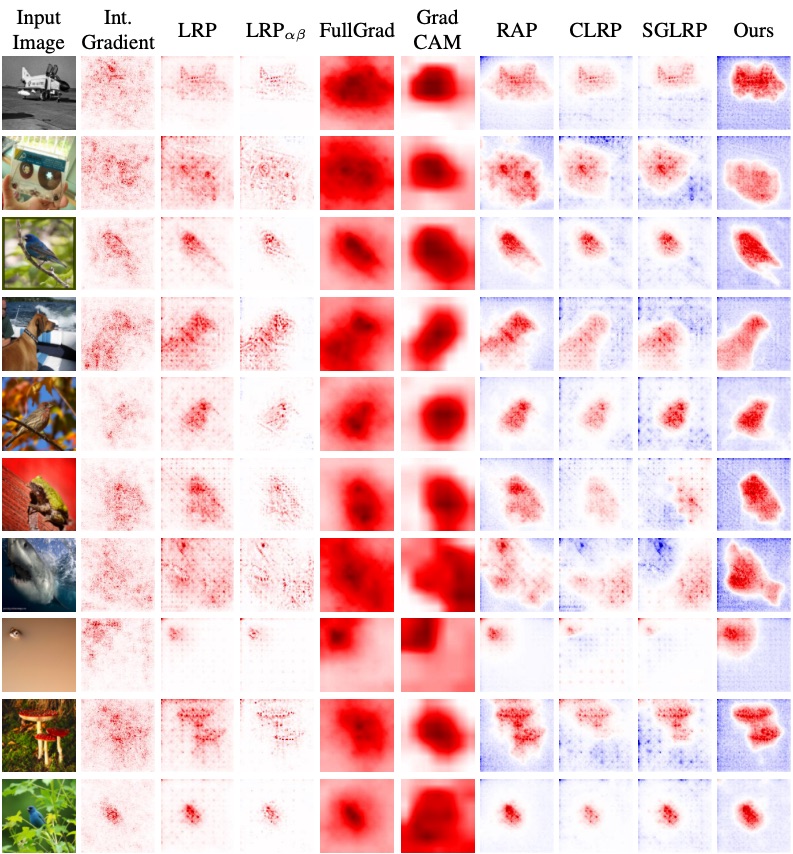}
    \caption{Visualization of the top predicted class of a ResNet-50 ImageNet trained network.}
    \label{fig:top_class_12}
\end{figure}
\clearpage
\begin{figure}
    \centering
    \includegraphics[width=\textwidth]{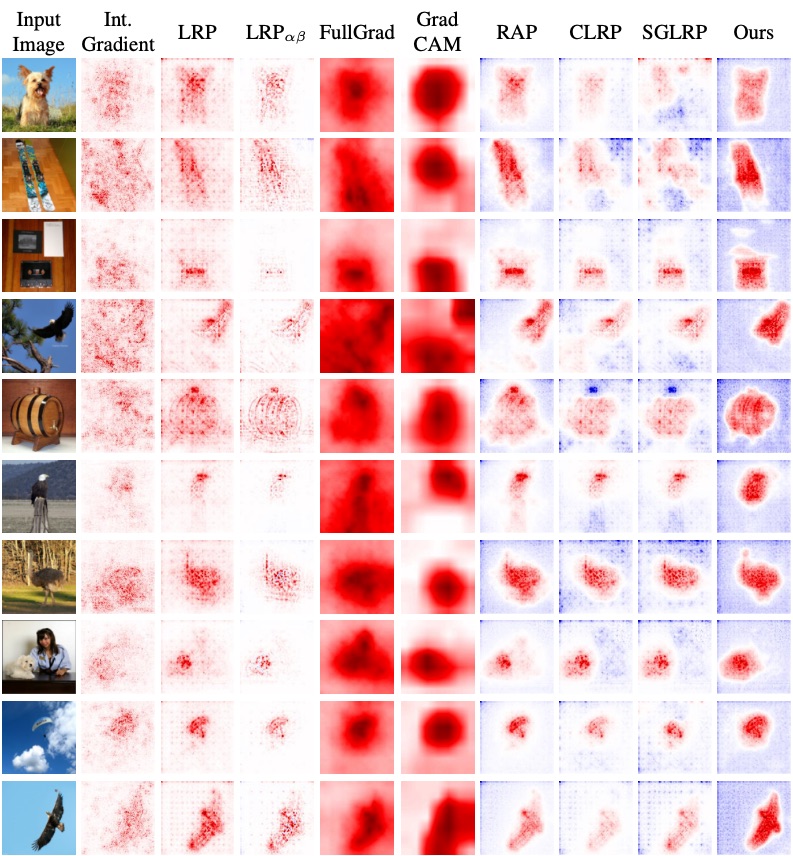}
    \caption{Visualization of the top predicted class of a ResNet-50 ImageNet trained network.}
    \label{fig:top_class_13}
\end{figure}
\clearpage
\begin{figure}
    \centering
    \includegraphics[width=\textwidth]{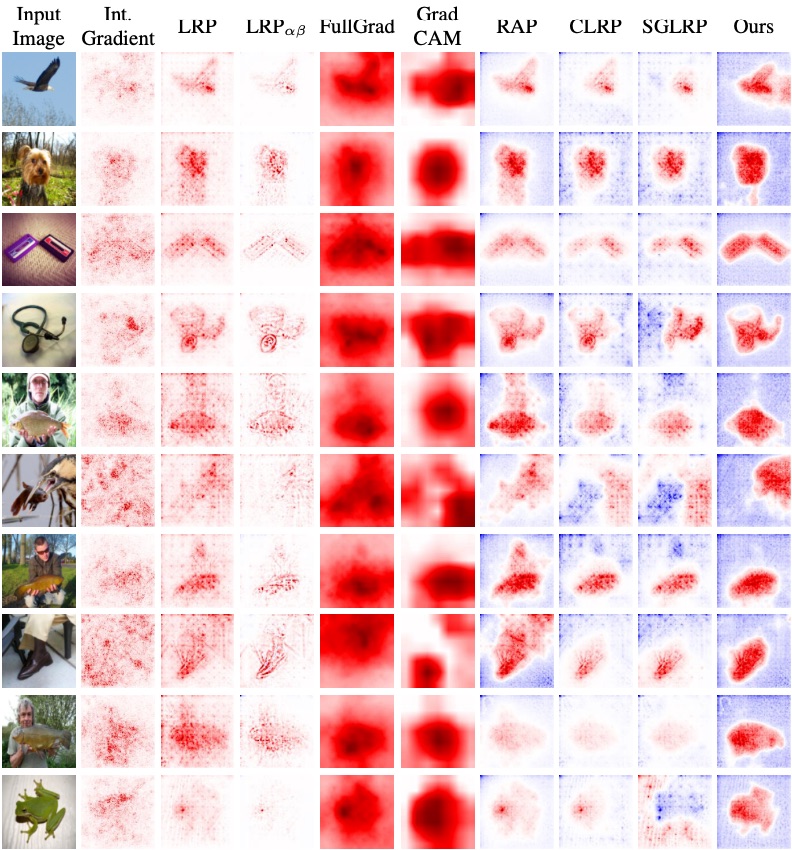}
    \caption{Visualization of the top predicted class of a ResNet-50 ImageNet trained network.}
    \label{fig:top_class_14}
\end{figure}

\clearpage
\begin{figure}
    \centering
    \includegraphics[width=\textwidth]{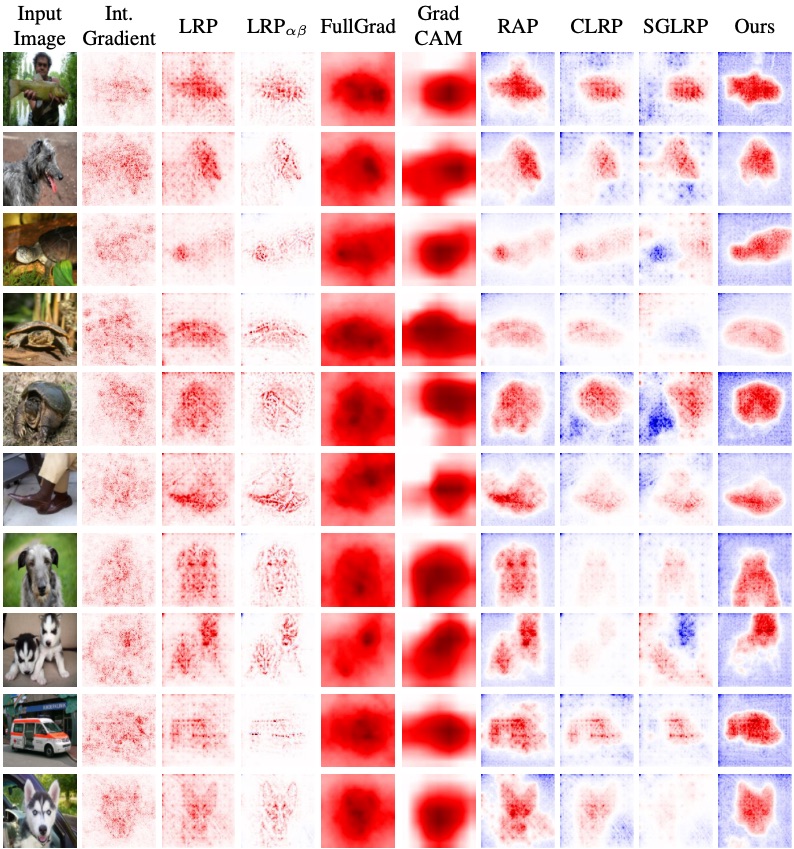}
    \caption{Visualization of the top predicted class of a ResNet-50 ImageNet trained network.}
    \label{fig:top_class_15}
\end{figure}

\clearpage
\section{Additional Results - AlexNet Probes}
\begin{figure}[h]
    \centering
    \begin{tabular}{cccccc}
         Input & \multirow{2}{*}{Layer 1.} & \multirow{2}{*}{Layer 2.} & \multirow{2}{*}{Layer 3.} & \multirow{2}{*}{Layer 4.} & \multirow{2}{*}{Layer 5.}  \\
     Image &  &  & & &   \\
        \includegraphics[width=0.1\textwidth]{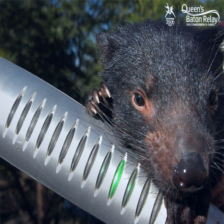} &
        \includegraphics[width=0.1\textwidth]{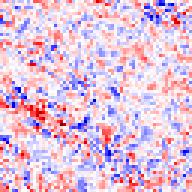} &
        \includegraphics[width=0.1\textwidth]{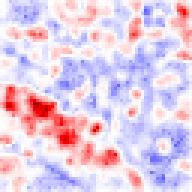} &
        \includegraphics[width=0.1\textwidth]{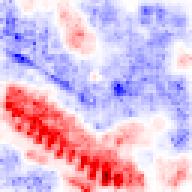} &
        \includegraphics[width=0.1\textwidth]{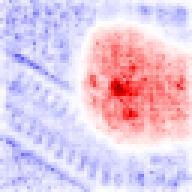} &
        \includegraphics[width=0.1\textwidth]{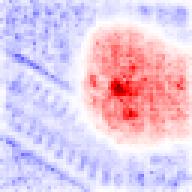}   \\
        \includegraphics[width=0.1\textwidth]{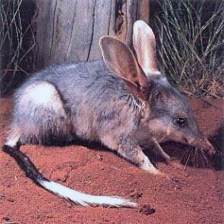} &
        \includegraphics[width=0.1\textwidth]{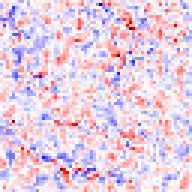} &
        \includegraphics[width=0.1\textwidth]{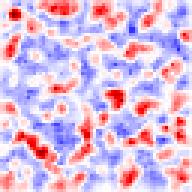} &
        \includegraphics[width=0.1\textwidth]{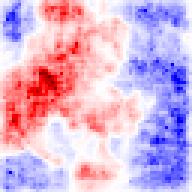} &
        \includegraphics[width=0.1\textwidth]{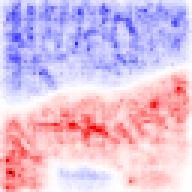} &
        \includegraphics[width=0.1\textwidth]{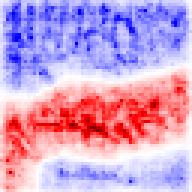}   \\  
        \includegraphics[width=0.1\textwidth]{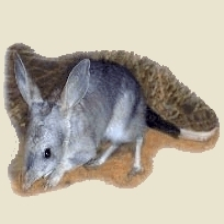} &
        \includegraphics[width=0.1\textwidth]{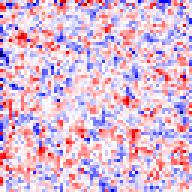} &
        \includegraphics[width=0.1\textwidth]{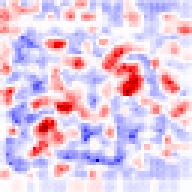} &
        \includegraphics[width=0.1\textwidth]{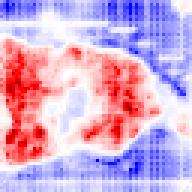} &
        \includegraphics[width=0.1\textwidth]{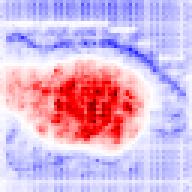} &
        \includegraphics[width=0.1\textwidth]{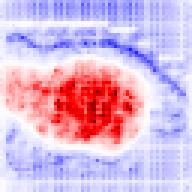}   \\
        \includegraphics[width=0.1\textwidth]{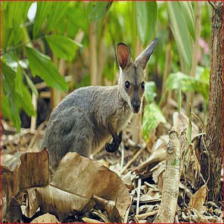} &
        \includegraphics[width=0.1\textwidth]{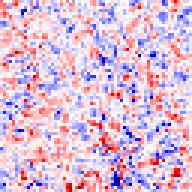} &
        \includegraphics[width=0.1\textwidth]{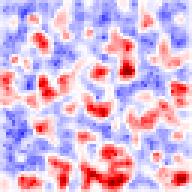} &
        \includegraphics[width=0.1\textwidth]{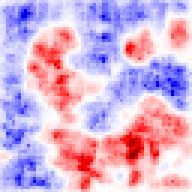} &
        \includegraphics[width=0.1\textwidth]{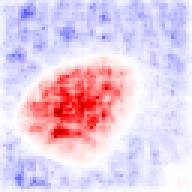} &
        \includegraphics[width=0.1\textwidth]{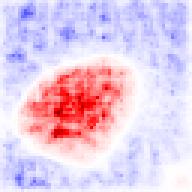}   \\  
        \includegraphics[width=0.1\textwidth]{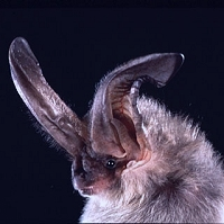} &
        \includegraphics[width=0.1\textwidth]{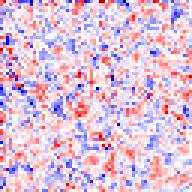} &
        \includegraphics[width=0.1\textwidth]{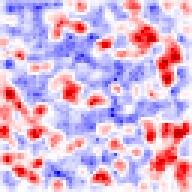} &
        \includegraphics[width=0.1\textwidth]{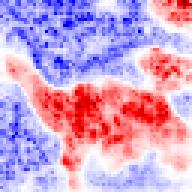} &
        \includegraphics[width=0.1\textwidth]{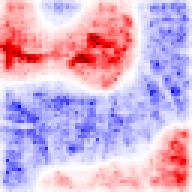} &
        \includegraphics[width=0.1\textwidth]{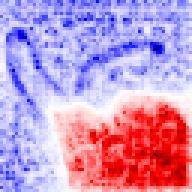}   \\  
        \includegraphics[width=0.1\textwidth]{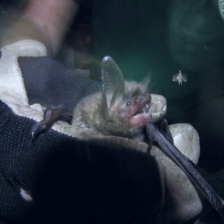} &
        \includegraphics[width=0.1\textwidth]{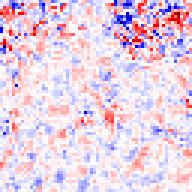} &
        \includegraphics[width=0.1\textwidth]{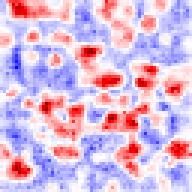} &
        \includegraphics[width=0.1\textwidth]{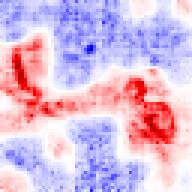} &
        \includegraphics[width=0.1\textwidth]{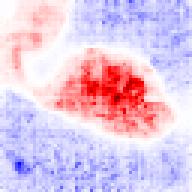} &
        \includegraphics[width=0.1\textwidth]{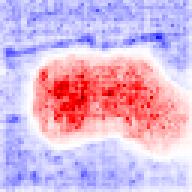}   \\ 
        \includegraphics[width=0.1\textwidth]{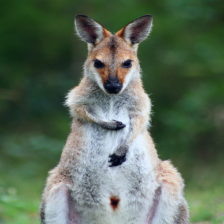} &
        \includegraphics[width=0.1\textwidth]{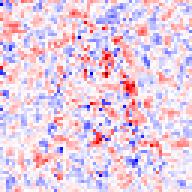} &
        \includegraphics[width=0.1\textwidth]{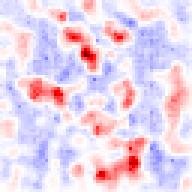} &
        \includegraphics[width=0.1\textwidth]{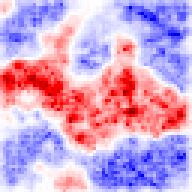} &
        \includegraphics[width=0.1\textwidth]{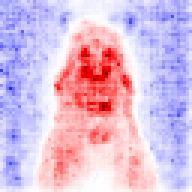} &
        \includegraphics[width=0.1\textwidth]{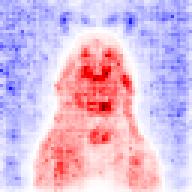}   \\ 
        \includegraphics[width=0.1\textwidth]{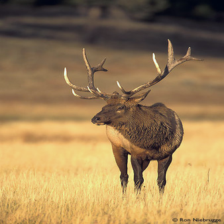} &
        \includegraphics[width=0.1\textwidth]{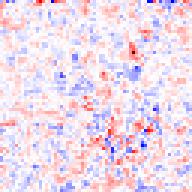} &
        \includegraphics[width=0.1\textwidth]{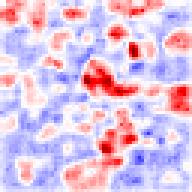} &
        \includegraphics[width=0.1\textwidth]{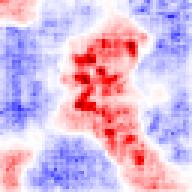} &
        \includegraphics[width=0.1\textwidth]{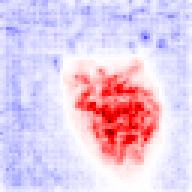} &
        \includegraphics[width=0.1\textwidth]{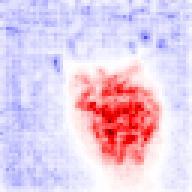}   \\ 
    \end{tabular}
    \caption{The visualization after each layer using linear probes on the RotNet AlexNet model.}
    \label{fig:top_class_15}
\end{figure}

\clearpage
\section{Additional Results - Self-labeling (SeLa) Method}

For each explainability method, we present results obtained by simply projecting the self labeling label with the highest probability (a very simplified version of our procedure that does not involve the nearest neighbor computation) as well as the one for Alg.~\ref{alg:ssl}.

\begin{figure}[h]
    \centering
    \begin{tabular}{ccccccccc}
         Input & \multirow{2}{*}{LRP$_{\alpha \beta}$} & LRP$_{\alpha \beta}$ & \multirow{2}{*}{LRP} & LRP & \multirow{2}{*}{RAP} & RAP & \multirow{2}{*}{Ours} & Ours  \\
     Image &  & with Alg.\ref{alg:ssl} &  & with Alg.\ref{alg:ssl} &&with Alg.\ref{alg:ssl} & & with Alg.\ref{alg:ssl}   \\
        \includegraphics[width=0.09\textwidth]{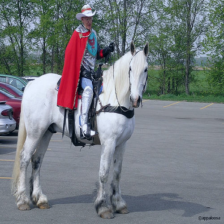} &
        \includegraphics[width=0.09\textwidth]{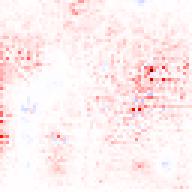} &
        \includegraphics[width=0.09\textwidth]{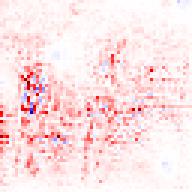} &
        \includegraphics[width=0.09\textwidth]{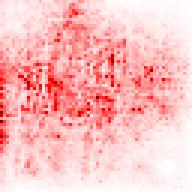} &
        \includegraphics[width=0.09\textwidth]{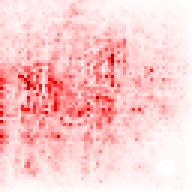} &
        \includegraphics[width=0.09\textwidth]{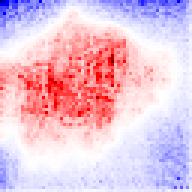} &
        \includegraphics[width=0.09\textwidth]{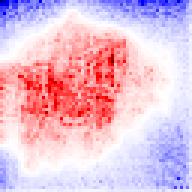} &
        \includegraphics[width=0.09\textwidth]{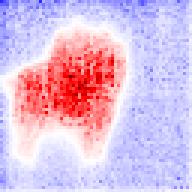} &
        \includegraphics[width=0.09\textwidth]{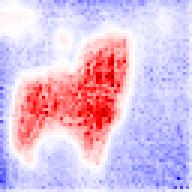}   \\
        \includegraphics[width=0.09\textwidth]{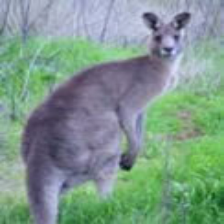} &
        \includegraphics[width=0.09\textwidth]{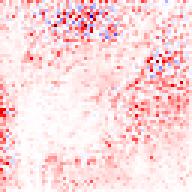} &
        \includegraphics[width=0.09\textwidth]{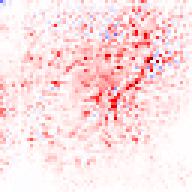} &
        \includegraphics[width=0.09\textwidth]{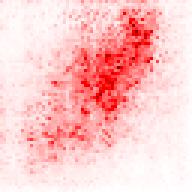}   &
        \includegraphics[width=0.09\textwidth]{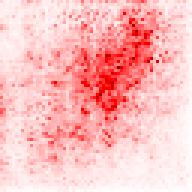}&
        \includegraphics[width=0.09\textwidth]{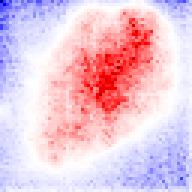} &
        \includegraphics[width=0.09\textwidth]{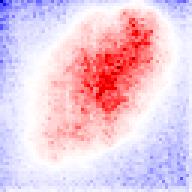} &
        \includegraphics[width=0.09\textwidth]{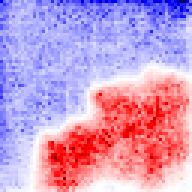} &
        \includegraphics[width=0.09\textwidth]{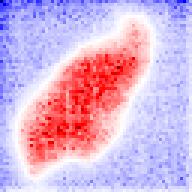}   \\
        \includegraphics[width=0.09\textwidth]{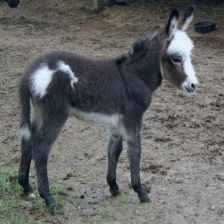} &
        \includegraphics[width=0.09\textwidth]{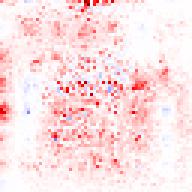} &
        \includegraphics[width=0.09\textwidth]{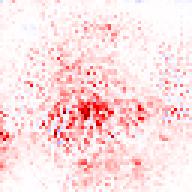}          &
        \includegraphics[width=0.09\textwidth]{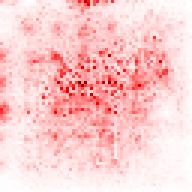}   &
        \includegraphics[width=0.09\textwidth]{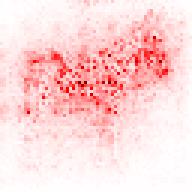}&
        \includegraphics[width=0.09\textwidth]{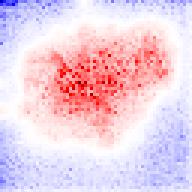} &
        \includegraphics[width=0.09\textwidth]{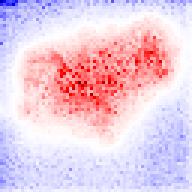} &
        \includegraphics[width=0.09\textwidth]{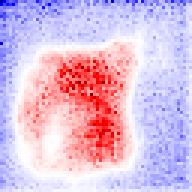} &
        \includegraphics[width=0.09\textwidth]{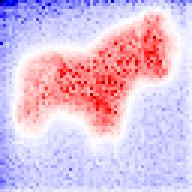}   \\
        \includegraphics[width=0.09\textwidth]{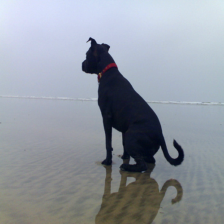} &
        \includegraphics[width=0.09\textwidth]{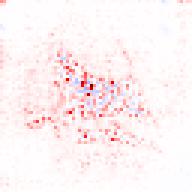}   &
        \includegraphics[width=0.09\textwidth]{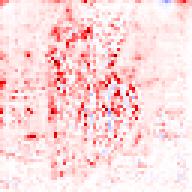}  &
        \includegraphics[width=0.09\textwidth]{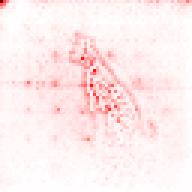}  &
        \includegraphics[width=0.09\textwidth]{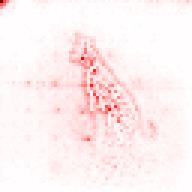} &
        \includegraphics[width=0.09\textwidth]{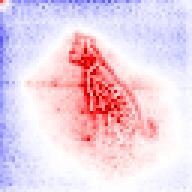} &
        \includegraphics[width=0.09\textwidth]{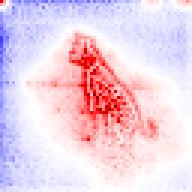} &
        \includegraphics[width=0.09\textwidth]{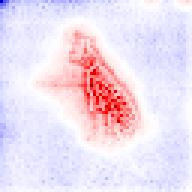} &
        \includegraphics[width=0.09\textwidth]{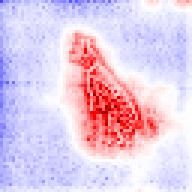}   \\
        \includegraphics[width=0.09\textwidth]{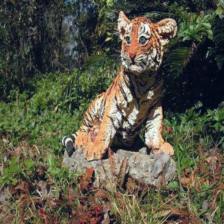} &
        \includegraphics[width=0.09\textwidth]{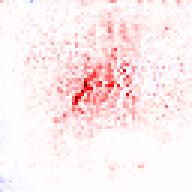} &
        \includegraphics[width=0.09\textwidth]{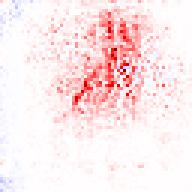} &
        \includegraphics[width=0.09\textwidth]{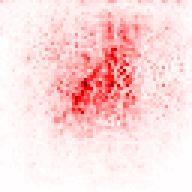} &
        \includegraphics[width=0.09\textwidth]{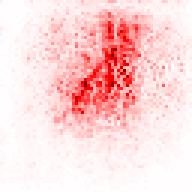}  &
        \includegraphics[width=0.09\textwidth]{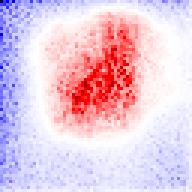} &
        \includegraphics[width=0.09\textwidth]{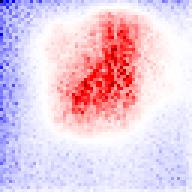} &
        \includegraphics[width=0.09\textwidth]{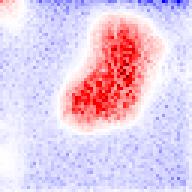} &
        \includegraphics[width=0.09\textwidth]{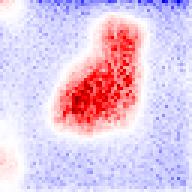}   \\
        \includegraphics[width=0.09\textwidth]{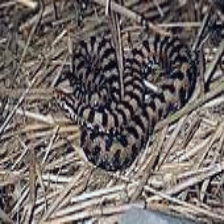} &
        \includegraphics[width=0.09\textwidth]{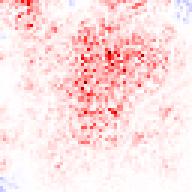} &
        \includegraphics[width=0.09\textwidth]{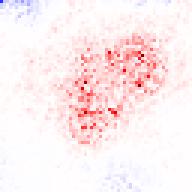} &
        \includegraphics[width=0.09\textwidth]{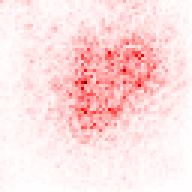}  &
        \includegraphics[width=0.09\textwidth]{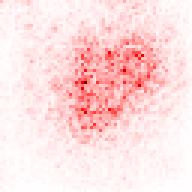} &
        \includegraphics[width=0.09\textwidth]{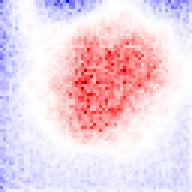} &
        \includegraphics[width=0.09\textwidth]{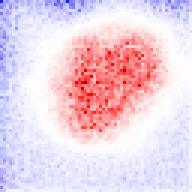} &
        \includegraphics[width=0.09\textwidth]{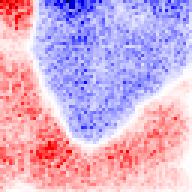} &
        \includegraphics[width=0.09\textwidth]{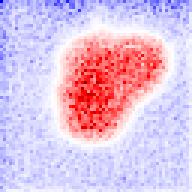}   \\
        \includegraphics[width=0.09\textwidth]{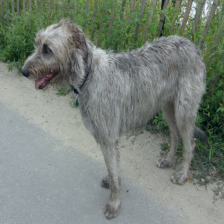} &
        \includegraphics[width=0.09\textwidth]{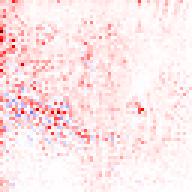} &
        \includegraphics[width=0.09\textwidth]{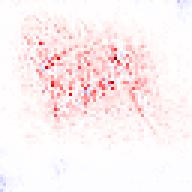}  &
        \includegraphics[width=0.09\textwidth]{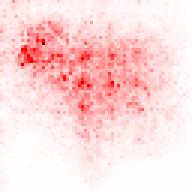}      &
        \includegraphics[width=0.09\textwidth]{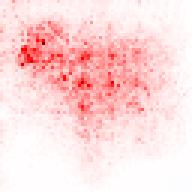}  &
        \includegraphics[width=0.09\textwidth]{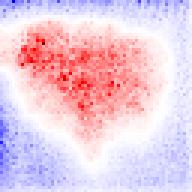} &
        \includegraphics[width=0.09\textwidth]{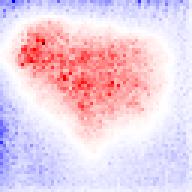} &
        \includegraphics[width=0.09\textwidth]{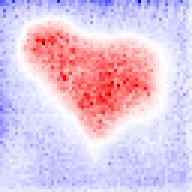} &
        \includegraphics[width=0.09\textwidth]{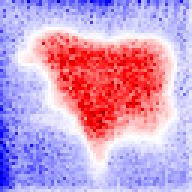}   \\
        \includegraphics[width=0.09\textwidth]{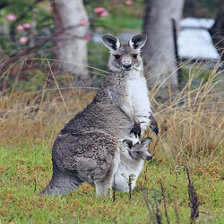} &
        \includegraphics[width=0.09\textwidth]{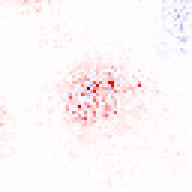}  &
        \includegraphics[width=0.09\textwidth]{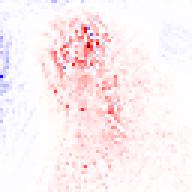} &
        \includegraphics[width=0.09\textwidth]{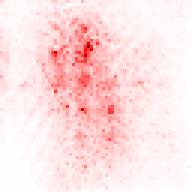}  &
        \includegraphics[width=0.09\textwidth]{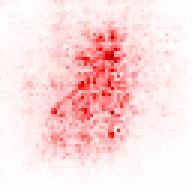} &
        \includegraphics[width=0.09\textwidth]{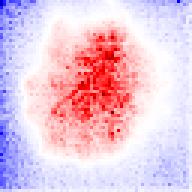} &
        \includegraphics[width=0.09\textwidth]{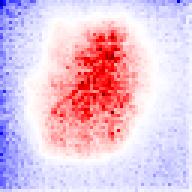} &
        \includegraphics[width=0.09\textwidth]{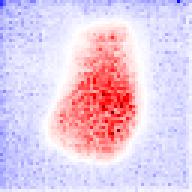} &
        \includegraphics[width=0.09\textwidth]{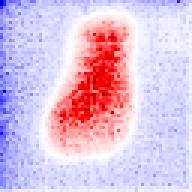}   \\

    \end{tabular}
    \caption{The visualization of different explanability methods on Resnet-50 trained in self-supervised regime using self-labeling method.}
    \label{fig:top_class_15}
\end{figure}
\clearpage
\section{Additional Results - SCAN  Method}
For each explainability method, we present results obtained by simply projecting the self labeling label with the highest probability (a very simplified version of our procedure that does not involve the nearest neighbor computation) as well as the one for Alg.~\ref{alg:ssl}.

\begin{figure}[h]
    \centering
    \begin{tabular}{ccccccccc}
         Input & \multirow{2}{*}{LRP$_{\alpha \beta}$} & LRP$_{\alpha \beta}$ & \multirow{2}{*}{LRP} & LRP & \multirow{2}{*}{RAP} & RAP & \multirow{2}{*}{Ours} & Ours  \\
     Image &  & with Alg.\ref{alg:ssl} &  & with Alg.\ref{alg:ssl} &&with Alg.\ref{alg:ssl} & & with Alg.\ref{alg:ssl}  \\
        \includegraphics[width=0.09\textwidth]{figures/supp/ssl/1650_input.png} &
        \includegraphics[width=0.09\textwidth]{figures/supp/ssl/selfl_lrp_ab/heatmap_1650.jpg}&
        \includegraphics[width=0.09\textwidth]{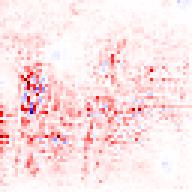}&
        \includegraphics[width=0.09\textwidth]{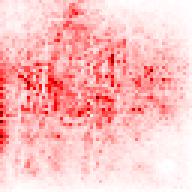}&
        \includegraphics[width=0.09\textwidth]{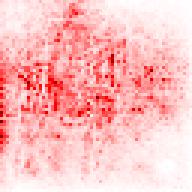} &
        \includegraphics[width=0.09\textwidth]{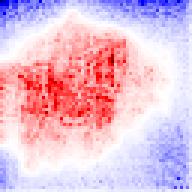} &
        \includegraphics[width=0.09\textwidth]{figures/supp/ssl/selfl_rap_1nn/heatmap_1650.jpg}&
        \includegraphics[width=0.09\textwidth]{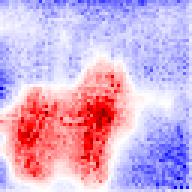} &
        \includegraphics[width=0.09\textwidth]{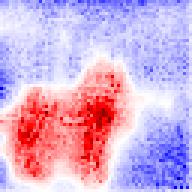}\\
        \includegraphics[width=0.09\textwidth]{figures/supp/ssl/141_input.png} &
        \includegraphics[width=0.09\textwidth]{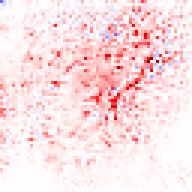}     &
        \includegraphics[width=0.09\textwidth]{figures/supp/ssl/selfl_lrp_ab_1nn/heatmap_141.jpg}     &
        \includegraphics[width=0.09\textwidth]{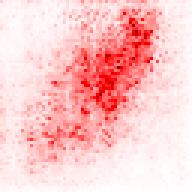} &
        \includegraphics[width=0.09\textwidth]{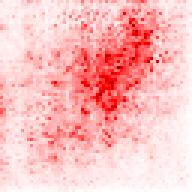}&
        \includegraphics[width=0.09\textwidth]{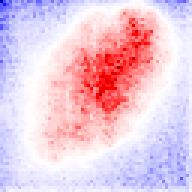}&
        \includegraphics[width=0.09\textwidth]{figures/supp/ssl/selfl_rap_1nn/heatmap_141.jpg}&
        \includegraphics[width=0.09\textwidth]{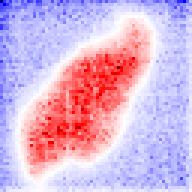} &
        \includegraphics[width=0.09\textwidth]{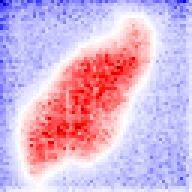} \\
        \includegraphics[width=0.09\textwidth]{figures/supp/ssl/1719_input.png} &
        \includegraphics[width=0.09\textwidth]{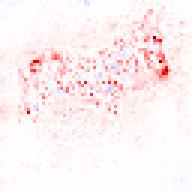}&
        \includegraphics[width=0.09\textwidth]{figures/supp/ssl/lrp_nnab_scan/heatmap_1719.jpg} &
        \includegraphics[width=0.09\textwidth]{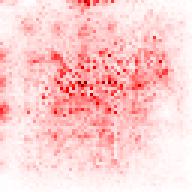}&
        \includegraphics[width=0.09\textwidth]{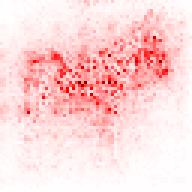}&
        \includegraphics[width=0.09\textwidth]{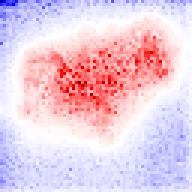}&
        \includegraphics[width=0.09\textwidth]{figures/supp/ssl/selfl_rap_1nn/heatmap_1719.jpg}&
        \includegraphics[width=0.09\textwidth]{figures/supp/ssl/selfl_agf_1nn/heatmap_1719.jpg} &
        \includegraphics[width=0.09\textwidth]{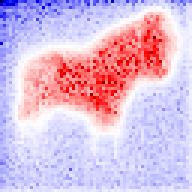}   \\
        \includegraphics[width=0.09\textwidth]{figures/supp/ssl/546_input.png} &
        \includegraphics[width=0.09\textwidth]{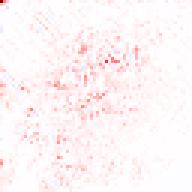} &
        \includegraphics[width=0.09\textwidth]{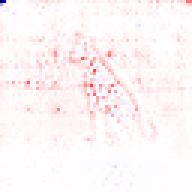}&
        \includegraphics[width=0.09\textwidth]{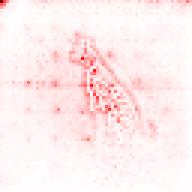}&
        \includegraphics[width=0.09\textwidth]{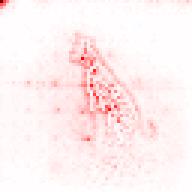}&
        \includegraphics[width=0.09\textwidth]{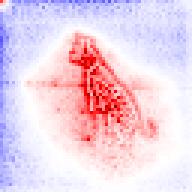}  &
        \includegraphics[width=0.09\textwidth]{figures/supp/ssl/selfl_rap_1nn/heatmap_546.jpg}&
        \includegraphics[width=0.09\textwidth]{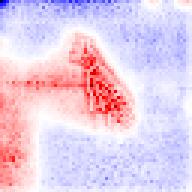}&
        \includegraphics[width=0.09\textwidth]{figures/supp/ssl/selfl_agf_1nn/heatmap_546.jpg}\\
        \includegraphics[width=0.09\textwidth]{figures/supp/ssl/22_input.png} &
        \includegraphics[width=0.09\textwidth]{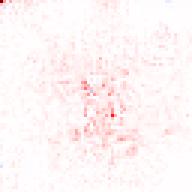}  &
        \includegraphics[width=0.09\textwidth]{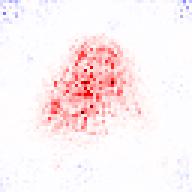}&
        \includegraphics[width=0.09\textwidth]{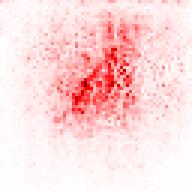}&
        \includegraphics[width=0.09\textwidth]{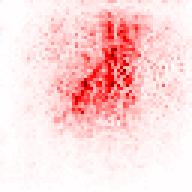}&
        \includegraphics[width=0.09\textwidth]{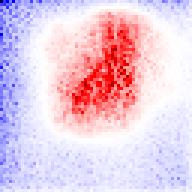}  &
        \includegraphics[width=0.09\textwidth]{figures/supp/ssl/selfl_rap_1nn/heatmap_22.jpg}  &
        \includegraphics[width=0.09\textwidth]{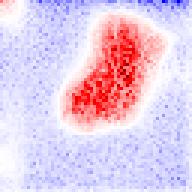} &
        \includegraphics[width=0.09\textwidth]{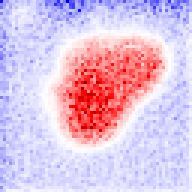} \\
        \includegraphics[width=0.09\textwidth]{figures/supp/ssl/31_input.png} &
        \includegraphics[width=0.09\textwidth]{figures/supp/ssl/selfl_lrp_ab_1nn/heatmap_22.jpg}&
        \includegraphics[width=0.09\textwidth]{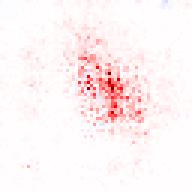}&
        \includegraphics[width=0.09\textwidth]{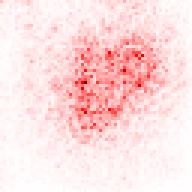}  &
        \includegraphics[width=0.09\textwidth]{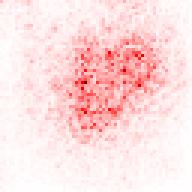}&
        \includegraphics[width=0.09\textwidth]{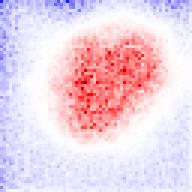} &
        \includegraphics[width=0.09\textwidth]{figures/supp/ssl/selfl_rap_1nn/heatmap_31.jpg}&
        \includegraphics[width=0.09\textwidth]{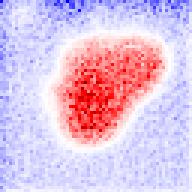} &
        \includegraphics[width=0.09\textwidth]{figures/supp/ssl/scan_agf_1nn/heatmap_31.jpg}\\
        \includegraphics[width=0.09\textwidth]{figures/supp/ssl/490_input.png} &
        \includegraphics[width=0.09\textwidth]{figures/supp/ssl/selfl_lrp_ab/heatmap_490.jpg} &
        \includegraphics[width=0.09\textwidth]{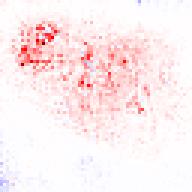} &
        \includegraphics[width=0.09\textwidth]{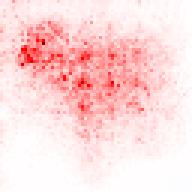} &
        \includegraphics[width=0.09\textwidth]{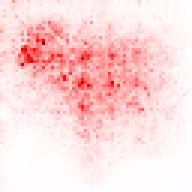}&
        \includegraphics[width=0.09\textwidth]{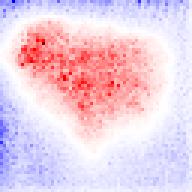} &
        \includegraphics[width=0.09\textwidth]{figures/supp/ssl/selfl_rap_1nn/heatmap_490.jpg}&
        \includegraphics[width=0.09\textwidth]{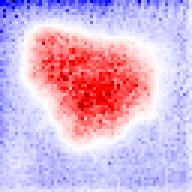}&
        \includegraphics[width=0.09\textwidth]{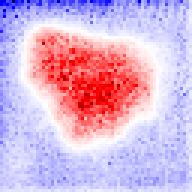} \\
        \includegraphics[width=0.09\textwidth]{figures/supp/ssl/266_input.png} &
        \includegraphics[width=0.09\textwidth]{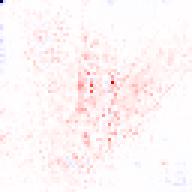}&
        \includegraphics[width=0.09\textwidth]{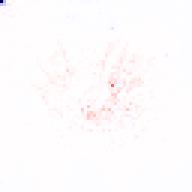}&
        \includegraphics[width=0.09\textwidth]{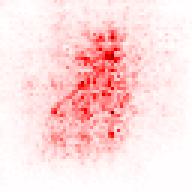}&
        \includegraphics[width=0.09\textwidth]{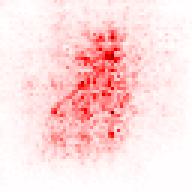}&
        \includegraphics[width=0.09\textwidth]{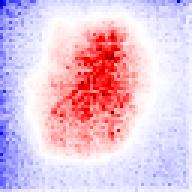} &
        \includegraphics[width=0.09\textwidth]{figures/supp/ssl/scan_rap/heatmap_266.jpg} &
        \includegraphics[width=0.09\textwidth]{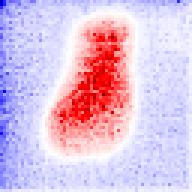} &
        \includegraphics[width=0.09\textwidth]{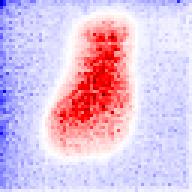}          \\

    \end{tabular}
    \caption{The visualization of different explanability methods on Resnet-50 trained in self-supervised regime using SCAN method.}
    \label{fig:top_class_15}
\end{figure}
\end{document}